\documentclass{article}
\pdfoutput=1

\usepackage{arxiv}

\usepackage[utf8]{inputenc} 
\usepackage[T1]{fontenc}    
\usepackage[hypertexnames=false]{hyperref}       
\usepackage{url}            
\usepackage{booktabs}       
\usepackage{amsfonts}       
\usepackage{nicefrac}       
\usepackage{microtype}      

\usepackage{lipsum}         
\usepackage{graphicx}
\usepackage{natbib}
\usepackage{doi}

\usepackage{mathtools,amsthm,amssymb}
\usepackage{algorithm}
\usepackage{algpseudocodex}
\usepackage{xcolor}
\usepackage{etoolbox}
\usepackage{todonotes}
\usepackage{environ}
\usepackage{bm}
\usepackage{enumitem}
\allowdisplaybreaks[4]

\usepackage{cleveref}       
\usepackage{crossreftools}
\hypersetup{
  colorlinks=true,
  linkcolor=red!75!black,
  citecolor=blue!50!black
}

\title{Understanding Gradient Orthogonalization for Deep Learning via Non-Euclidean Trust-Region Optimization}
\renewcommand{\shorttitle}{Understanding Gradient Orthogonalization via Non-Euclidean Trust-Region Optimization}

\newif\ifuniqueAffiliation
\uniqueAffiliationtrue

\ifuniqueAffiliation 
\author{
  Dmitry Kovalev \\
  Yandex Research \\
  \texttt{dakovalev1@gmail.com}
}
\else
\usepackage{authblk}

\author[1]{%
  Dmitry Kovalev\thanks{\texttt{dakovalev1@gmail.com}}%
}
\affil[1]{Yandex Research}
\fi

\newtheorem{theorem}{Theorem}
\newtheorem{lemma}{Lemma}
\newtheorem{corollary}{Corollary}

\crefname{assumption}{assumption}{assumptions}

\crefname{problem}{problem}{problems}
\creflabelformat{problem}{(#2#1#3)}

\newcounter{aequation}
\NewEnviron{aequation}{\refstepcounter{aequation}$$\BODY\eqno{\text{(A\theaequation)}}$$}
\crefname{aequation}{assumption}{assumptions}
\creflabelformat{aequation}{(#2A#1#3)}

\DeclarePairedDelimiter{\norm}{\|}{\|}
\DeclarePairedDelimiter{\sqn}{\|}{\|^2}

\DeclarePairedDelimiter{\infnorm}{\|}{\|_{\infty}}

\def\<#1,#2>{\langle #1,#2\rangle}

\DeclareMathOperator{\dom}{dom}

\DeclareMathOperator{\diam}{diam}

\DeclareMathOperator{\prox}{prox}

\DeclareMathOperator{\ri}{ri}
\DeclareMathOperator{\closure}{cl}

\DeclareMathOperator*{\argmin}{argmin}

\newcommand{\E}[1]{\mathbb{E}\left[#1\right]}

\newcommand{\Ed}[2]{\mathbb{E}_{#1}\left[#2\right]}

\newcommand{\R}{\mathbb{R}}

\newcommand{\cA}{\mathcal{A}}

\newcommand{\cD}{\mathcal{D}}

\newcommand{\cL}{\mathcal{L}}

\newcommand{\cO}{\mathcal{O}}

\newcommand{\cU}{\mathcal{U}}

\newcommand{\cX}{\mathcal{X}}

\newcommand{\mG}{\mathbf{G}}
\newcommand{\mH}{\mathbf{H}}

\newcommand{\mM}{\mathbf{M}}

\newcommand{\mO}{\mathbf{O}}

\newcommand{\mU}{\mathbf{U}}
\newcommand{\mV}{\mathbf{V}}

\newcommand{\mX}{\mathbf{X}}

\makeatletter
\newcommand{\annotatehypertarget}[1]{\Hy@raisedlink{\hypertarget{#1}{}}}
\makeatother
\newcounter{annotatecount}
\newcounter{annotateidx}
\newcounter{annotatejdx}
\newcounter{annotatelabelcount}
\newcounter{annotateglobalindex}
\newcommand{\atran}[2]{\stepcounter{annotatecount}\overset{\text{\annotatehypertarget{\alph{annotatecount}\theannotateglobalindex}{(\hyperlink{desc\alph{annotatecount}\theannotateglobalindex}{\alph{annotatecount}})}}}{#1}\csgdef{annotatedescription\theannotatecount}{#2}}
\newcommand{\aeq}[1]{\atran{=}{#1}}
\newcommand{\aleq}[1]{\atran{\leq}{#1}}
\newcommand{\ageq}[1]{\atran{\geq}{#1}}

\newcommand{\annotateinitused}{\setcounter{annotateidx}{0}\whileboolexpr{test{\ifnumless{\theannotateidx}{\theannotatecount}}}{\stepcounter{annotateidx}\csgdef{aused\theannotateidx}{0}}}
\newcommand{\annotategetlabels}{\setcounter{annotatejdx}{0}\setcounter{annotatelabelcount}{0}\whileboolexpr{test{\ifnumless{\theannotatejdx}{\theannotatecount}}}{\stepcounter{annotatejdx}\ifcsequal{annotatedescription\theannotateidx}{annotatedescription\theannotatejdx}{\csgdef{aused\theannotatejdx}{1}\stepcounter{annotatelabelcount}\csedef{annotatelabel\theannotatelabelcount}{\alph{annotatejdx}}}{}}}
\newcommand{\annotateprintlabels}{\setcounter{annotatejdx}{0}\whileboolexpr{test{\ifnumless{\theannotatejdx}{\theannotatelabelcount}}}{\stepcounter{annotatejdx}\ifnumequal{\theannotatejdx}{\theannotatelabelcount}{\ifnumequal{\theannotatejdx}{1}{}{~and~}}{}\annotatehypertarget{desc\csuse{annotatelabel\theannotatejdx}\theannotateglobalindex}{(\hyperlink{\csuse{annotatelabel\theannotatejdx}\theannotateglobalindex}{\csuse{annotatelabel\theannotatejdx}})}\ifnumless{\theannotatejdx}{\theannotatelabelcount}{\ifnumless{\theannotatejdx+1}{\theannotatelabelcount}{,~}{}}{}}}
\newcommand{\annotate}{\annotateinitused\setcounter{annotateidx}{0}\whileboolexpr{test{\ifnumless{\theannotateidx}{\theannotatecount}}}{\stepcounter{annotateidx}\ifcsstring{aused\theannotateidx}{0}{\ifnumequal{\theannotateidx}{1}{}{;~}\annotategetlabels\annotateprintlabels~\csuse{annotatedescription\theannotateidx}}{}}\setcounter{annotatecount}{0}\stepcounter{annotateglobalindex}}

\newcommand{\sX}{\cX}
\newcommand{\sM}{\R^{m\times n}}

\newcommand{\mind}[1]{\hspace{#1}&\hspace{-#1}}
\newcommand{\alertR}[1]{{\color{red}#1}}

\newcommand{\subgrad}{\hat{\nabla}}
\newcommand{\mSigma}{\mathbf{\Sigma}}
\newcommand{\msigma}{\boldsymbol{\sigma}}
\newcommand{\ox}{\overline{x}}

\DeclarePairedDelimiter{\sqne}{\|}{\|_2^2}
\DeclarePairedDelimiter{\norme}{\|}{\|_2}
\DeclarePairedDelimiter{\sqns}{\|}{\|_{\mathrm{op}}^2}
\DeclarePairedDelimiter{\norms}{\|}{\|_{\mathrm{op}}}

\DeclarePairedDelimiter{\normn}{\|}{\|_{\mathrm{nuc}}}
\DeclareMathOperator{\orth}{orth}

\undef{\diam}
\DeclareMathOperator{\diam}{diam_{\norm{\cdot}}}
\undef{\epsilon}
\newcommand{\epsilon}{\varepsilon}

\begin{document}
\maketitle

\begin{abstract}
  Optimization with matrix gradient orthogonalization has recently demonstrated impressive results in the training of deep neural networks \citep{jordan2024muon,liu2025muon}. In this paper, we provide a theoretical analysis of this approach. In particular, we show that the orthogonalized gradient method can be seen as a first-order trust-region optimization method, where the trust-region is defined in terms of the matrix spectral norm. Motivated by this observation, we develop the stochastic non-Euclidean trust-region gradient method with momentum, which recovers the Muon optimizer \citep{jordan2024muon} as a special case, along with normalized SGD and signSGD with momentum \citep{cutkosky2020momentum,sun2023momentum}. In addition, we prove state-of-the-art convergence results for the proposed algorithm in a range of scenarios, which involve arbitrary non-Euclidean norms, constrained and composite problems, and non-convex, star-convex, first- and second-order smooth functions. Finally, our theoretical findings provide an explanation for several practical observations, including the practical superiority of Muon compared to the Orthogonal-SGDM algorithm of \citet{tuddenham2022orthogonalising} and the importance of weight decay in the training of large-scale language models.
\end{abstract}


\section{Introduction}

Over the past couple of decades, a substantial amount of optimization research has been dedicated to adaptive gradient optimization algorithms \citep{duchi2011adaptive,tieleman2012lecture,kingma2014adam,gupta2018shampoo,reddi2019convergence}, with the primary application being the training of deep neural networks \citep{lecun2015deep}. One of the most notable results of this line of research is the development of the AdamW optimizer \citep{loshchilov2017decoupled,zhuang2022understanding}, which has become the standard algorithm of choice for training large language models (LLMs) \citep{achiam2023gpt,liu2024deepseek,grattafiori2024llama,team2023gemini}. However, recently, \citet{jordan2024muon,liu2025muon} have made significant progress in the ambitious task of surpassing AdamW in training LLMs using the idea of neural network optimization with orthogonalized gradients \citep{tuddenham2022orthogonalising,jordan2024muon}. In our paper, we aim to establish theoretical foundations for this promising research direction.
Formally speaking, we consider the following composite optimization problem:
\begin{equation}\label[problem]{eq:main}
  \min_{x \in \sX} \left[F(x) = f(x) + R(x)\right],
\end{equation}
where $\sX$ is a finite-dimensional vector space endowed with the inner product $\<\cdot,\cdot>\colon \sX\times \sX \to \R$, $f(\cdot)\colon \sX \to \R$ is a bounded from below and differentiable objective function, and $R(\cdot)\colon \sX \to \R \cup \{+\infty\}$ is a proper,\footnote{Function $R(x)$ is called proper if there exists $x \in \sX$ such that $R(x)$ is finite.} closed, and convex regularizer. Our goal is to study the convergence properties of gradient methods for solving \cref{eq:main} in the {\em stochastic non-Euclidean smooth} setting. We provide a formal description of this setting below and justify our interest in this setting in the upcoming \Cref{sec:motivation,sec:summary}.

{\bf Stochastic gradient estimator.}
We assume access to a stochastic estimator $g(\cdot\,;\xi)\colon \sX \to \sX$ of the gradient $\nabla f(\cdot)$, where $\xi \sim \cD$ is a random variable sampled from a probability distribution $\cD$. We assume that the stochastic gradient estimator $g(\cdot;\xi)$ is unbiased and has bounded variance, that is, the following relations hold:
\begin{aequation}\label{eq:variance}
  \Ed{\xi \sim \cD}{g(x;\xi)} = \nabla f(x)
  \quad\text{and}\quad
  \Ed{\xi \sim \cD}{\sqne{g(x;\xi) - \nabla f(x)}} \leq \sigma^2
  \quad\text{for all}\;
  x \in \sX,
\end{aequation}
where $\sigma > 0$ is a positive variance parameter, and $\norme{\cdot}$ is the standard Euclidean norm induced by the inner product $\<\cdot,\cdot>$, i.e., $\norme{x} = \sqrt{\<x,x>}$. These assumptions have been widely adopted for the analysis of many stochastic gradient optimization algorithms \citep{ghadimi2013stochastic,ghadimi2016accelerated,cutkosky2020momentum,sun2023momentum,horvath2023stochastic,gorbunov2020linearly}.

{\bf Non-Euclidean norm setting and Lipschitz continuous gradient.}
We assume that vector space $\sX$ is equipped with a norm $\norm{\cdot}\colon \sX \to \R_+$, which possibly does not coincide with the Euclidean norm $\norme{\cdot}$. In addition, we assume that the gradient $\nabla f(\cdot)$ is Lipschitz continuous with respect to the norm $\norm{\cdot}$, that is, the following inequality holds:
\begin{aequation}\label{eq:L}
  \norm{\nabla f(x) - \nabla f(x')}_* \leq L \norm{x-x'}
  \quad\text{for all}\;
  x,x' \in \sX,
\end{aequation}
where $L > 0$ is the gradient Lipschitz constant, and $\norm{\cdot}_*\colon \sX \to \R_+$ is the dual norm associated with $\norm{\cdot}$, i.e., $\norm{x}_* = \sup_{\norm{x'}\leq 1} \<x,x'>$ for all $x \in \sX$. The assumption of gradient Lipschitz continuity is also widespread in the analysis of first-order optimization methods \citep{ghadimi2013stochastic,gower2019sgd,cutkosky2020momentum,horvath2023stochastic,gorbunov2020linearly}. It is important to highlight that while \Cref{eq:L} uses the dual norm $\norm{\cdot}_*$ to measure the difference between the gradients, the variance in \Cref{eq:variance} is measured with respect to the Euclidean norm $\sqne{\cdot}$, which is necessary to properly utilize the unbiasedness property of the stochastic gradient estimator $g(\cdot;\xi)$. Therefore, we need to provide a connection between these norms using the following inequality:
\begin{aequation}\label{eq:norm}
  \norm{x}_* \leq \rho\cdot\norme{x}
  \quad\text{for all}\;
  x \in \sX,
\end{aequation}
where $\rho > 0$ is a positive constant. Note that such a constant always exists due to the norm equivalence theorem, which always holds in the finite-dimensional space $\sX$.

\subsection{Main Motivation: Optimization with Orthogonalized Gradients}\label{sec:motivation}

The main motivation for this work is the recently proposed idea of using {\em orthogonalized gradient updates} for the training of deep neural networks \citep{tuddenham2022orthogonalising,jordan2024muon}. To explain the idea, we consider the problem of minimizing a differentiable matrix function $F(\cdot)\colon \sM \to \R$
\begin{equation}\label[problem]{eq:mat}
  \min_{\mX \in \sM} F(\mX),
\end{equation}
which is clearly a special instance of the main \cref{eq:main}, as long as we choose $\sX$ to be the space of $m\times n$ matrices, $\sX = \sM$, and assume zero regularization, $R \equiv 0$. The iterations of the simplest version of the orthogonalized gradient method for solving \cref{eq:mat} can be written as follows:
\begin{equation}\label{eq:ogd}
  \mX_{k+1} = \mX_k - \eta \mO_k,\quad
  \mO_k = \orth(\mG_k),\quad
  \mG_k = \nabla F(\mX_k).
\end{equation}
where $\eta > 0$ is the stepsize, and $\orth(\cdot) \colon \sM \to \sM$ is the orthogonalization function, which is defined as follows:
\begin{equation}
  \orth(\mG) = (\mG\mG^\top)^{\frac{\dagger}{2}}\mG\quad\text{for all}\quad \mG \in \sM,
\end{equation}
where $\frac{\dagger}{2}$ denotes the square root of the pseudoinverse matrix. The name <<orthogonalized>> is attributed to the fact that in the case where the gradient $\nabla F(\mX_k)$ is a square and full-rank matrix, the update $\mO_k$ can be equivalently expressed as $\mO_k = \mU_k \mV_k^\top$, where $\mU_k$ and $\mV_k$ are the orthogonal matrix parts of the singular value decomposition of the gradient, i.e., $\nabla F(\mX_k) = \mU_k \mSigma_k \mV_k^\top$, where $\mSigma_k \in \sM$ is a diagonal matrix with non-negative entries.

Combining iterations~\eqref{eq:ogd} with momentum leads to the Muon optimizer \citep{jordan2024muon} for solving the stochastic version of \cref{eq:mat}. It has been empirically shown that this algorithm can considerably outperform AdamW in the training of both small-scale \citep{jordan2024muon} and large-scale \citep{liu2025muon} language models. However, to the best of our knowledge, {\em the understanding of the effectiveness of Muon is highly limited}. In addition, it is worth mentioning the Orthogonal-SGDM optimizer \citep{tuddenham2022orthogonalising}, which differs from Muon only in the order in which the momentum and orthogonalization are applied. Unfortunately, \citet{tuddenham2022orthogonalising} were not able to outperform a well-tuned standard SGD with momentum using their algorithm, and \citet{jordan2024muon} reported that Muon performs substantially better in practice than Orthogonal-SGDM. To the best of our knowledge, {\em an explanation for this phenomenon is missing from the literature}.

\subsection{Summary of Contributions and Related Work}\label{sec:summary}

Motivated by the above discussion, we set the main goal of this paper to develop the theoretical foundations for optimization with orthogonalized gradients. More specifically, we provide the key contributions listed below.\footnote{In \Cref{sec:FW} we compare our results with the concurrent work of \citet{pethick2025training}, which we found only after the initial version of our paper appeared online.}

\textbf{Gradient orthogonalization as non-Euclidean trust-region optimization.}
\citet{jordan2024muon,bernstein2024old} found an interpretation of the orthogonalized gradient method~\eqref{eq:ogd} as the standard gradient descent under the non-Euclidean {\em matrix spectral norm}. However, this interpretation is highly inaccurate, as further discussed in \Cref{sec:ogd}. In \Cref{sec:tr}, we develop a {\em completely new interpretation} of the gradient orthogonalization as a {\em trust-region gradient method}, where the trust-region is defined in terms of the non-Euclidean spectral norm. This interpretation is accurate because it allows us to recover the iterations~\eqref{eq:ogd} exactly. Moreover, using this interpretation, we obtain several meaningful theoretical results, which we describe further. We also discuss the benefits of using the matrix spectral norm in the training of neural networks in \Cref{sec:example}.

\textbf{Stochastic non-Euclidean trust-region gradient method with momentum.} Motivated by our new interpretation of gradient orthogonalization, in \Cref{sec:str}, we develop the {\em stochastic non-Euclidean trust-region gradient method with momentum} (\Cref{alg}) for solving \cref{eq:main}. This algorithm is designed to work with an arbitrary non-Euclidean norm $\norm{\cdot}$, and recovers several existing optimization algorithms: normalized SGD with momentum \citep{cutkosky2020momentum}, signSGD with momentum \citep{sun2023momentum}, and Muon. We also prove the $1/\epsilon^4$ iteration complexity of \Cref{alg} for non-convex functions, which yields {\em the first convergence result for Muon} and matches the existing optimal results for SGD-type methods for non-convex functions \citep{ghadimi2013stochastic,arjevani2023lower}. In addition, in \Cref{sec:osgdm}, we use our results to provide a theoretical explanation of why Muon outperforms its natural contender, Orthogonal-SGDM \citep{tuddenham2022orthogonalising}, in practice, and justify the subtle differences in the design of these algorithms.

\textbf{Convergence analysis for star-convex functions.}
In \Cref{sec:alg_decay}, we develop the {\em stochastic non-Euclidean trust-region gradient method with weight decay} (\Cref{alg:decay}) for solving \cref{eq:main} by utilizing the popular weight decay mechanism \citep{loshchilov2017decoupled}. We obtain the improved $1/\epsilon^3$ iteration complexity of \Cref{alg:decay} under the star-convexity assumption \citep{nesterov2006cubic}. In addition, in \Cref{sec:D}, we prove the same $1/\epsilon^3$ complexity result for \Cref{alg} without weight decay, by additionally assuming the boundedness of $\dom R$. Our results support the practical evidence for the importance of incorporating weight decay into Muon for the training of large-scale language models, and explain why it may be less important for the training of small-scale language models, as reported by \citet{liu2025muon,jordan2024muon}. Another important remark is that our theoretical results for star-convex functions hold substantial practical interest because there is empirical evidence \citep{zhou2019sgd,kleinberg2018alternative} suggesting that deep neural networks may adhere to star-convexity or its variants.

\textbf{Convergence analysis for second-order smooth functions.}
Inspired by the theoretical results of \citet{cutkosky2020momentum}, in \Cref{sec:hess}, we develop the {\em stochastic non-Euclidean trust-region gradient method with extrapolation} for solving \cref{eq:main} by incorporating a certain extrapolation step into \Cref{alg:decay}. We obtain the $1/\epsilon^{3.5}$ complexity for non-convex functions under the additional second-order smoothness assumption, which matches the results of \citet{cutkosky2020momentum,sun2023momentum} for normalized SGD and signSGD and improves upon the standard $1/\epsilon^4$ complexity for SGD-type methods for non-convex functions \citep{ghadimi2013stochastic,arjevani2023lower}. In addition, we provide a convergence analysis for \Cref{alg:hess} for star-convex functions, which also shows an improvement over the results for \Cref{alg,alg:decay}.

\section{Non-Euclidean Trust-Region Gradient Method}\label{sec:tr}

\subsection{Gradient Descent as Majorization-Minimization}\label{sec:mm}
It is generally known that the standard proximal gradient descent can be seen as an instance of the majorization-minimization (MM) algorithm \citep{hunter2004tutorial}. In particular, we can replace the objective function $F(x)$ in \cref{eq:main} with the following approximation at an arbitrary point $z \in \sX$:
\begin{equation}\label{eq:major}
  \cU(x;z) = f(z) + \<\nabla f(z),x-z> + \tfrac{1}{2\theta}\sqn{x-z} + R(x),
\end{equation}
which is constructed by replacing function $f(x)$ with its first-order Taylor approximation at point $z$ and adding the squared norm regularization $\tfrac{1}{2\theta}\sqn{x-z}$. This approximation is accurate at point $z$, i.e., $F(z) = \cU(z;z)$, and using standard arguments, one can show that $\cU(x;z)$ majorizes $F(x)$, i.e., $F(x) \leq \cU(x;z)$ for all $x \in \sX$, as long as $\theta \geq 1/L$ and \Cref{eq:L} holds. Consequently, the iterations of the MM algorithm can be written as follows:
\begin{equation}\label{eq:mm}
  x_{k+1} = \argmin_{x \in \sX} \cU(x;x_k).
\end{equation}
One can immediately observe that these iterations coincide with the proximal gradient method \citep{nesterov2013gradient} as long as we assume the Euclidean norm setting, i.e., $\norm{\cdot} = \norm{\cdot}_* = \norme{\cdot}$:
\begin{equation}\label{eq:prox_gd}
  x_{k+1} = \prox_{\theta R(\cdot)}(x_k - \theta \nabla f(x_k)).
\end{equation}
The convergence guarantees for this algorithm were established in the Euclidean setting for both non-convex and convex functions $f(x)$ by \citet{nesterov2013gradient}.

\subsection{Non-Euclidean Trust-Region Gradient Method}

The trust-region optimization approach \citep{conn2000trust,jiang2023beyond} is a notable alternative to the MM principle discussed above. It is usually employed when access to the second-order derivatives of the objective function is assumed. This approach involves minimizing a second-order Taylor-like approximation of the objective function at a given point within a certain neighborhood of that point, which is called the trust-region.

Further, we construct a first-order, i.e., gradient, trust-region optimization algorithm for solving \cref{eq:main}. More specifically, we replace the objective function $F(x)$ with the following approximation at an arbitrary point $z \in \sX$:
\begin{equation}\label{eq:approx}
  \cA(x;z) = f(z) + \<\nabla f(z),x-z> + R(x),
\end{equation}
which does not necessarily majorize function $F(x)$ due to the lack of squared norm regularization, in contrast to $\cU(x;z)$ in \cref{eq:major}. Thus, the iterations of the trust-region method minimize this approximation only within the ball of radius $\eta > 0$ defined in terms of the non-Euclidean norm $\norm{\cdot}$ as follows:
\begin{equation}\label{eq:tr}
  x_{k+1} = \argmin_{x\in\sX} \cA(x;x_k)
  \quad\text{s.t.}\quad
  \norm{x-x_k}\leq \eta.
\end{equation}
As a particular example, it is easy to verify that in the Euclidean norm setting with zero regularization, i.e., where $\norm{\cdot} = \norm{\cdot}_* = \norme{\cdot}$ and $R \equiv 0$, these iterations coincide with normalized gradient descent \citep{shor2012minimization}:
\begin{equation}\label{eq:ngd}
  x_{k+1} = x_k -  \frac{\eta}{\norme{\nabla f(x_k)}}\nabla f(x_k).
\end{equation}
The convergence guarantees for the iterations~\eqref{eq:ngd} were established in the convex case by \cite{grimmer2019convergence} and in some non-convex scenarios by \citet{nesterov1984minimization}. However, to the best of our knowledge, convergence results for the trust-region gradient method~\eqref{eq:tr} with a general non-Euclidean norm $\norm{\cdot}$ and an arbitrary proper closed and convex regularizer $R(\cdot)$ are unknown for both convex and non-convex functions $f(x)$. We provide such results for non-convex functions in \Cref{sec:tr_nonconvex} and for star-convex functions in \Cref{sec:D}.

\subsection{Gradient Orthogonalization for Matrix Function Optimization}\label{sec:ogd}

Now, we revisit the minimization \cref{eq:mat} over the space of matrices $\sX = \sM$, which is a special instance of the main \cref{eq:main} with zero regularization $R\equiv 0$. We choose the non-Euclidean norms $\norm{\cdot}$ and $\norm{\cdot}_*$ to be the spectral $\norms{\cdot}$ and the nuclear $\normn{\cdot}$ matrix norms, respectively. That is, for an arbitrary matrix $\mX \in \sM$, we have $\norms{\mX} = \infnorm{\msigma(\mX)}$ and $\normn{\mX} = \norm{\msigma(\mX)}_1$, where $\msigma(\mX)$ is the vector of singular values of $\mX$. Thus, one can verify that {\em the non-Euclidean trust-region gradient method~\eqref{eq:tr} is exactly equivalent to the orthogonalized gradient method~\eqref{eq:ogd}}. This is one of the key observations of our work. It is in sharp contrast with \citet{bernstein2024old}, who suggested the interpretation of the orthogonalized gradient method as the gradient descent~\eqref{eq:mm} in the same non-Euclidean setting. As mentioned in \Cref{sec:summary}, this interpretation is highly inaccurate because the iterations~\eqref{eq:mm} reduce to the following:
\begin{equation}\label{eq:ogd_wrong}
  \mX_{k+1} = \mX_k - \theta\normn{\mG_k}\mO_k,
\end{equation}
where $\mG_k = \nabla F(\mX_k)$, and $\mO_k \in \sM$ is defined in \cref{eq:ogd}. Indeed, there is an extra nuclear norm factor $\normn{\mG_k}$ in the update~\eqref{eq:ogd_wrong}. More importantly, in contrast to this interpretation of \citet{bernstein2024old}, we can use our trust-region approach to establish convergence guarantees for Muon. 

\subsection{Convergence Analysis for Non-Convex Functions}\label{sec:tr_nonconvex}

In this section, we provide a convergence analysis for the non-Euclidean trust-region gradient method~\eqref{eq:tr} for solving the non-stochastic version of \cref{eq:main} with the non-convex objective function $f(x)$ in the following \Cref{thm:tr_nonconvex} and \Cref{cor:tr_nonconvex}. The proof is available in \Cref{proof:tr_nonconvex}.
\begin{theorem}\label{thm:tr_nonconvex}
  Let \Cref{eq:L} hold and let $x_0 \in \dom R$. Then the iterations~\eqref{eq:tr} satisfy the following inequality:
  \begin{equation}
    \min_{k=1,\ldots,K} \norm{\nabla f(x_k) + \subgrad R_k}_* \leq \frac{\Delta_0}{\eta K} + \frac{3L\eta}{2},
  \end{equation}
  where $\subgrad R_k \in \partial R(x_k)$, $\Delta_0 = F(x_0) - \inf_x F(x)$.
\end{theorem}
\begin{corollary}\label{cor:tr_nonconvex}
  To reach the precision $\min_{k=1\ldots K} \norm{\nabla f(x_k) + \subgrad R_k}_* \leq \epsilon$ by the iterations~\eqref{eq:tr} under the conditions of \Cref{thm:tr_nonconvex}, it is sufficient to choose the stepsize $\eta$ and the number of iterations $K$ as follows:
  \begin{equation}
    \eta = \cO\left(\frac{\epsilon}{L}\right),\qquad K = \cO\left(\frac{L\Delta_0}{\epsilon^2}\right).
  \end{equation}
\end{corollary}
We make two remarks regarding \Cref{thm:tr_nonconvex} and \Cref{cor:tr_nonconvex}. First, the generalized stationarity condition $\norm{\nabla f(x_k) + \subgrad R_k}_* \leq \epsilon$ is widely adopted in the literature \citep{nesterov2013gradient}. In the case of zero regularization, $R\equiv 0$, it reduces to the standard definition of an $\epsilon$-stationary point, $\norm{\nabla f(x_k)}_* \leq \epsilon$. Second, the obtained iteration complexity $K = \cO\left({L\Delta_0}/{\epsilon^2}\right)$ matches the standard result for gradient descent in the Euclidean setting by \citet{nesterov2013gradient} and cannot be improved in general \citep{carmon2020lower}.

\section{Stochastic Non-Euclidean Trust-Region Gradient Method with Momentum}\label{sec:str}

\subsection{The Algorithm}\label{sec:alg}

In this section, we present \Cref{alg} for solving \cref{eq:main} in the stochastic setting. We call this algorithm the stochastic non-Euclidean trust-region gradient method with momentum. The main idea is to replace the gradient $\nabla f(x_k)$ in the trust-region gradient method~\eqref{eq:tr} with the momentum term $m_{k+1}$, which is updated according to \cref{eq:momentum}. While the idea is inspired by the results for normalized SGD with momentum by \citet{cutkosky2020momentum}, their analysis is not applicable in the case of a non-Euclidean norm $\norm{\cdot}$ and a nonzero regularizer $R \not\equiv 0$. We provide the convergence analysis of \Cref{alg} for non-convex functions in \Cref{sec:str_nonconvex} and for star-convex functions in \Cref{sec:D}.

It is also important to highlight that by choosing different norms $\norm{\cdot}$, \Cref{alg} can be reduced to several important special instances: {\bf(i)} Euclidean norm, $\norm{\cdot} = \norme{\cdot}$ -- normalized SGD with momentum \citep{cutkosky2020momentum}; {\bf(ii)} infinity-norm, $\norm{\cdot} = \infnorm{\cdot}$ -- signSGD with momentum \citep{sun2023momentum}; {\bf(iii)} spectral norm, $\norm{\cdot} = \norms{\cdot}$ -- Muon \citep{jordan2024muon}. The latter reduction is implied by the discussion in \Cref{sec:ogd}.

\begin{algorithm}[t]
  \caption{Stochastic Non-Euclidean Trust-Region Gradient Method with Momentum}
  \label{alg}
  \begin{algorithmic}[1]
    \State {\bf input:} $x_0, m_0\in \sX$
    \State {\bf parameters:} stepsize $\eta > 0$, momentum $\alpha \in (0,1)$, number of iterations $K \in \{1,2,\ldots\}$
    \For{$k=0,1,\ldots, K-1$}
    \State Sample $\xi_k \sim \cD$
    \State Compute $m_{k+1}$ as follows:
    \begin{align}
      \label{eq:momentum}
      m_{k+1} &= (1-\alpha)m_k + \alpha g(x_k;\xi_k)
    \end{align}
    \State Compute $x_{k+1}$ as follows:
    \begin{align}
      \label{eq:str}
      x_{k+1} &= {\argmin_{x\in \sX}} [\<m_{k+1}, x> + R(x)]\quad\text{s.t.}\quad \norm{x-x_k} \leq \eta
    \end{align}
    \EndFor
    \State {\bf output:} $x_K \in \sX$
  \end{algorithmic}
\end{algorithm}
\subsection{Convergence Analysis for Non-Convex Functions}\label{sec:str_nonconvex}

In this section, we provide the convergence analysis for \Cref{alg} for solving \cref{eq:main} in the stochastic non-convex case in the following \Cref{thm:str_nonconvex}. The proof is available in \Cref{proof:str_nonconvex}.
\begin{theorem}\label{thm:str_nonconvex}
  Let \Cref{eq:L,eq:variance,eq:norm} hold, and let $x_0 \in \dom R$ and $m_0 = g(x_0,\xi_0)$. Then the iterations of \Cref{alg} satisfy the following inequality:
  \begin{equation}
    \E{\min_{k=1,\ldots,K}\norm{\nabla f(x_k) + \subgrad R_k}_*}
    \leq
    \frac{\Delta_0}{\eta K}
    +\frac{2\rho\sigma}{\alpha K}
    +2\sqrt{\alpha}\rho\sigma
    +\frac{7L\eta}{2}
    +\frac{2 L\eta}{\alpha},
  \end{equation}
  where $\subgrad R_k \in \partial R(x_k)$, $\Delta_0 = F(x_0) - \inf_x F(x)$.
\end{theorem}
\begin{corollary}\label{cor:str_nonconvex}
  To reach the precision $\E{\min_{k=1\ldots K} \norm{\nabla f(x_k) + \subgrad R_k}_*} \leq \epsilon$ by \Cref{alg} under the conditions of \Cref{thm:str_nonconvex}, it is sufficient to choose the parameters of \Cref{alg} as follows:
  \begin{align}
    \eta &= \cO\left(\min\left\{\frac{\epsilon}{L}, \frac{\epsilon^3}{\rho^2\sigma^2L}\right\}\right),
    \qquad
    \alpha = \cO\left(\min\left\{1, \frac{\epsilon^2}{\rho^2\sigma^2}\right\}\right),
    \\
    \label{eq:str_K_nonconvex}
    K &= \cO\left(\max\left\{
        \frac{\rho\sigma}{\epsilon},
        \frac{\rho^3\sigma^3}{\epsilon^3},
        \frac{L\Delta_0}{\epsilon^2},
        \frac{L\Delta_0\rho^2\sigma^2}{\epsilon^4}
    \right\}\right).
  \end{align}
\end{corollary}
Using \Cref{thm:str_nonconvex}, we obtain the explicit complexity result for \Cref{alg} in \Cref{cor:str_nonconvex}. Similarly to the convergence result in \Cref{cor:tr_nonconvex} for the deterministic non-Euclidean trust-region gradient method~\eqref{eq:tr}, \Cref{cor:str_nonconvex} establishes the convergence result for \Cref{alg} in terms of the generalized expected stationarity. The iteration complexity~\eqref{eq:str_K_nonconvex} is proportional to $1/\epsilon^{4}$. It matches the existing state-of-the-art results for SGD-type methods \citep{ghadimi2013stochastic,cutkosky2020momentum,sun2023momentum} and cannot be improved \citep{arjevani2023lower} under \Cref{eq:L,eq:variance,eq:norm}.

Another important remark is that \Cref{alg} with the spectral norm $\norm{\cdot} = \norms{\cdot}$ and zero regularizer $R \equiv 0$ exactly matches Muon. As previously discussed, there is another variant of the orthogonalized gradient method with momentum, namely, Orthogonal-SGDM. While Muon demonstrated strong results in the training of small-scale language models \citep{jordan2024muon}, it was reported by \citet{tuddenham2022orthogonalising,jordan2024muon} that Orthogonal-SGDM performs worse than Muon or a well-tuned SGD with momentum. Our theoretical results may provide a possible explanation for this practical difference. We discuss this in \Cref{sec:osgdm}.

\section{Algorithms with Weight Decay for Star-Convex Functions}\label{sec:star_cvx}
\newcommand{\decay}[1]{\alertR{#1}}

\subsection{Non-Euclidean Trust-Region Gradient Methods with Weight Decay}\label{sec:alg_decay}

In this section, we develop non-Euclidean trust-region gradient methods for solving \cref{eq:main} in the case where the objective function $f(x)$ is star-convex, that is, the following inequality holds:
\begin{aequation}\label{eq:star_cvx}
  f(\beta x^* + (1 - \beta)x) \leq \beta f(x^*) + (1-\beta)f(x)
  \quad\text{for all}\;x \in \sX,
\end{aequation}
where $x^* \in \sX$ is a solution to \cref{eq:main}.
We start with the non-stochastic version of the problem. It turns out that under \cref{eq:star_cvx}, it is possible to obtain improved convergence guarantees for the iterations~\eqref{eq:tr} as long as these iterations are bounded. Unfortunately, proving the boundedness of these iterations requires additional assumptions, such as a bounded $\dom R$, or bounded sublevel sets of the objective function $F(x)$, which often do not hold. We tackle this issue by shifting the center of the trust-region in \cref{eq:tr} by a factor of $\decay{(1-\beta)}$ towards the zero. This idea leads to the non-Euclidean trust-region gradient method with weight decay~\eqref{eq:tr_decay}.
\begin{equation}\label{eq:tr_decay}
  x_{k+1} = \argmin_{x\in\sX} \cA(x;x_k)
  \quad\text{s.t.}\quad
  \norm{x-\decay{(1-\beta)}x_k}\leq \eta.
\end{equation}
It is not hard to verify that the iterations~\eqref{eq:tr_decay} are bounded. On the other hand, the original and shifted trust-regions do not differ much. In particular, it is possible to show that they have a non-zero intersection. Hence, the shift modification will not hurt the convergence properties of the iterations~\eqref{eq:tr_decay}, which we prove in \Cref{sec:decay_convex}.

Furthermore, we apply a similar modification to \Cref{alg}. This modification leads to \Cref{alg:decay}, which we call the stochastic non-Euclidean trust-region gradient method with weight decay. Similar to the discussion in \Cref{sec:alg}, one can show that by choosing different norms $\norm{\cdot}$, \Cref{alg:decay} reduces to the normalized SGD, signSGD, and Muon with momentum and weight decay.

\begin{algorithm}[t]
  \caption{Stochastic Non-Euclidean Trust-Region Gradient Method with Weight Decay}
  \label{alg:decay}
  \begin{algorithmic}[1]
    \State {\bf input:} $x_0, m_0\in \sX$
    \State {\bf parameters:} $\eta > 0$, $\alpha \in (0,1)$, $K \in \{1,2,\ldots\}$, \decay{weight decay $\beta \in (0,1)$}
    \For{$k=0,1,\ldots, K-1$}
    \State Sample $\xi_k \sim \cD$
    \State Compute $m_{k+1}$ as follows:
    \begin{align}
      \label{eq:momentum_decay}
      m_{k+1} &= (1-\alpha)m_k + \alpha g(x_k;\xi_k)
    \end{align}
    \State Compute $x_{k+1}$ as follows:
    \begin{align}
      \label{eq:str_decay}
      x_{k+1} &= {\argmin_{x\in \sX}} [\<m_{k+1}, x> + R(x)]\quad\text{s.t.}\quad \norm{x-\decay{(1-\beta)}x_k} \leq \eta
    \end{align}
    \EndFor
    \State {\bf output:} $x_K \in \sX$
  \end{algorithmic}
\end{algorithm}

\subsection{Convergence Analysis for Star-Convex Functions}\label{sec:decay_convex}

In this section, we provide the convergence analysis for the iterations~\eqref{eq:tr_decay} and \Cref{alg:decay} for solving \cref{eq:main} in the star-convex case. We start with the result in \Cref{thm:tr_decay} for the deterministic iterations~\eqref{eq:tr_decay}. The proof is available in \Cref{proof:tr_decay}. Next, we obtain the convergence result for the stochastic \Cref{alg:decay} in \Cref{thm:str_decay}. The proof is available in \Cref{proof:str_decay}.
\begin{theorem}\label{thm:tr_decay}
  Let \Cref{eq:L,eq:star_cvx} hold, and let $x_0 \in \dom R$. Let the parameters $\eta$ and $\beta$ satisfy the following inequality:
  \begin{equation}\label{eq:eta_beta}
    \eta \geq \beta \max\left\{\norm{x_0}, \norm{x^*}\right\}.
  \end{equation}
  Then, the iterations~\eqref{eq:tr_decay} satisfy the following inequality:
  \begin{equation}
    F(x_K) - F(x^*) \leq (1-\beta)^K(F(x_0) - F(x^*)) + \frac{4L\eta^2}{\beta}.
  \end{equation}
\end{theorem}
\begin{theorem}\label{thm:str_decay}
  Let \Cref{eq:L,eq:variance,eq:norm,eq:star_cvx} hold, and let $x_0 \in \dom R$ and $m_0 = g(x_0,\xi_0)$. Let the parameters $\eta$ and $\beta$ satisfy \cref{eq:eta_beta}. Then, the output of \Cref{alg:decay} satisfies the following inequality:
  \begin{equation}
    \E{F(x_K) - F(x^*)} \leq (1-\beta)^K (F(x_0) - F(x^*)) + 2\eta\rho\sigma\left(\frac{1}{\alpha} + \frac{\sqrt{\alpha}}{\beta}\right) + \frac{4L\eta^2}{\beta}\left(1 + \frac{1}{\alpha}\right).
  \end{equation}
\end{theorem}
Finally, using \Cref{thm:tr_decay,thm:str_decay}, it is not hard to obtain the explicit iteration complexities for the iterations~\eqref{eq:tr_decay} and \Cref{alg:decay} in \Cref{cor:tr_decay,cor:str_decay}, respectively. The proofs are omitted due to their simplicity.
\begin{corollary}\label{cor:tr_decay}
  Under the conditions of \Cref{thm:tr_decay}, let $\norm{x_0} \leq \norm{x^*} = D$. To reach the precision $F(x_K) - F(x^*) \leq \epsilon$ by the iterations~\eqref{eq:tr_decay}, it is sufficient to choose the parameters $\eta$ and $\beta$ and the number of iterations $K$ as follows:
  \begin{equation}
    \eta = \beta D,\qquad
    \beta = \cO\left(\min\left\{1, \frac{\epsilon}{LD}\right\}\right),\qquad
    K = \tilde{\cO}\left(\max\left\{1,\frac{LD^2}{\epsilon}\right\}\right).
  \end{equation}
\end{corollary}
\begin{corollary}\label{cor:str_decay}
  Under the conditions of \Cref{thm:str_decay}, let $\norm{x_0} \leq \norm{x^*} = D$. To reach the precision $\E{F(x_K) - F(x^*)} \leq \epsilon$ by \Cref{alg:decay}, it is sufficient to choose the parameters as follows:
  \begin{align}
    \alpha &= \cO\left(\min\left\{1, \frac{\epsilon^2}{D^2\rho^2\sigma^2}\right\}\right),
    &
    \beta &= \cO\left(\min\left\{
        1,
        \frac{\epsilon}{LD^2},
        \frac{\epsilon}{D\rho\sigma},
        \frac{\epsilon^3}{D^3\rho^3\sigma^3},
        \frac{\epsilon^3}{LD^3\rho^2\sigma^2}
    \right\}\right),
    \\
    \eta &= \beta D,
    &
    K &= \tilde{\cO}\left(\max\left\{
        1,
        \frac{LD^2}{\epsilon},
        \frac{D\rho\sigma}{\epsilon},
        \frac{D^3\rho^3\sigma^3}{\epsilon^3},
        \frac{LD^3\rho^2\sigma^2}{\epsilon^3}
    \right\}\right).
  \end{align}
\end{corollary}

Both \Cref{cor:tr_decay,cor:str_decay} recover the standard iteration complexity $\cO(LD^2/\epsilon)$ of gradient descent \citep{nesterov2018lectures} for solving deterministic convex minimization problems, up to logarithmic factors. This complexity cannot be improved in the general non-Euclidean norm setting \citep{guzman2015lower}. In the stochastic setting, \Cref{cor:str_decay} implies complexity proportional to $1/\epsilon^3$. This improves upon the $1/\epsilon^{3.5}$ result of \citet{cutkosky2020momentum,sun2023momentum}, which was obtained only for normalized SGD and signSGD with momentum, and for non-convex second-order smooth functions.\footnote{Additional details on the correctness of this comparison are provided in \Cref{sec:cmp}.}

Another important remark is that \Cref{alg:decay} with the spectral norm $\norm{\cdot} = \norms{\cdot}$ and zero regularizer $R \equiv 0$ exactly matches Muon with weight decay, as previously discussed in \Cref{sec:alg_decay}. It was empirically shown by \citet{liu2025muon} that incorporating weight decay in Muon is crucial for outperforming AdamW in the training of large-scale language models. In particular, they reported that without weight decay, the iterations of Muon grow too large and the performance gains of Muon over AdamW diminish. Our theory suggests that combining Muon with weight decay may be the right solution for both issues. Indeed, we prove that the iterations of \Cref{alg:decay} are bounded and obtain the improved complexity $1/\epsilon^3$ of \Cref{alg:decay} in \Cref{cor:str_decay}.

\section{Algorithms with Extrapolation for Second-Order Smooth Functions}\label{sec:hess}
\newcommand{\hess}[1]{{\color{blue}#1}}

\subsection{Stochastic Non-Euclidean Trust-Region Gradient Method with Extrapolation}\label{sec:alg_hess}

In this section, we modify \Cref{alg,alg:decay} to solve \cref{eq:main} in the stochastic setting in the case where the objective function $f(x)$ has an $H$-Lipschitz Hessian with respect to the non-Euclidean norm $\norm{\cdot}$, which implies the following inequality:
\begin{aequation}\label{eq:H}
  \norm{(\nabla^2 f(x)- \nabla^2 f(x'))(x-x')}_* \leq H\sqn{x-x'}
  \quad\text{for all}\; x,x' \in \sX,
\end{aequation}
where $H>0$ is the Hessian Lipschitz constant. The idea for the modification is inspired by \citet{cutkosky2020momentum}. In particular, we replace the stochastic gradient $g(x_k,\xi_k)$ in the update of the momentum term $m_{k+1}$ in \cref{eq:momentum,eq:momentum_decay} with the stochastic gradient $g(\hess{\ox_k};\xi_k)$ computed at a different point $\hess{\ox_k}$, which is updated using the extrapolation step~\eqref{eq:ox_hess}. This modification leads to \Cref{alg:hess}, which we call the stochastic non-Euclidean trust-region gradient method with extrapolation. While we borrow the idea for the analysis of the extrapolation step~\eqref{eq:ox_hess} from \citet{cutkosky2020momentum,sun2023momentum}, their analysis is limited to normalized SGD and signSGD for non-convex functions with zero regularization $R\equiv 0$. In contrast, in \Cref{sec:hess_theory}, we provide a convergence analysis for \Cref{alg:hess} with an arbitrary non-Euclidean norm $\norm{\cdot}$ and regularizer $R(\cdot)$ in the non-convex and star-convex cases.

Note that, similarly to \Cref{alg:decay} described in \Cref{sec:alg_decay}, \Cref{alg:hess} makes use of weight decay in \cref{eq:str_hess} along with the extrapolation step~\eqref{eq:ox_hess} when dealing with star-convex functions. Another important remark is that \Cref{alg:hess} with the matrix spectral norm $\norms{\cdot}$ turns into a new variant of Muon with extrapolation. We discuss this further in \Cref{sec:hess_theory}.

\subsection{Convergence Analysis}\label{sec:hess_theory}

\begin{algorithm}[t]
  \caption{Stochastic Non-Euclidean Trust-Region Gradient Method with Extrapolation}
  \label{alg:hess}
  \begin{algorithmic}[1]
    \State {\bf input:} $x_0=\hess{\ox_0}, m_0\in \sX$
    \State {\bf parameters:} $\eta > 0$, $\alpha \in (0,1)$, $K \in \{1,2,\ldots\}$, \decay{$\beta \in (0,1)$}, \hess{extrapolation $\gamma > 0$}
    \For{$k=0,1,\ldots, K-1$}
    \State Sample $\xi_k \sim \cD$
    \State Compute $m_{k+1}$ as follows:
    \begin{align}
      \label{eq:momentum_hess}
      m_{k+1} &= (1-\alpha)m_k + \alpha g(\hess{\ox_k};\xi_k)
    \end{align}
    \State Compute $x_{k+1}$ as follows:
    \begin{align}
      \label{eq:str_hess}
      x_{k+1} &= {\argmin_{x\in \sX}} [\<m_{k+1}, x> + R(x)]\quad\text{s.t.}\quad \norm{x-\decay{(1-\beta)}x_k} \leq \eta
    \end{align}
    \State Compute $\hess{\ox_{k+1}}$ as follows:
    \begin{align}\label{eq:ox_hess}
      \hess{\ox_{k+1} = x^k + \gamma(x_{k+1} - x_k)}
    \end{align}
    \EndFor
    \State {\bf output:} $x_K \in \sX$
  \end{algorithmic}
\end{algorithm}

We provide the convergence analysis for \Cref{alg:hess} for solving \cref{eq:main} with the non-convex objective function in \Cref{thm:hess_nonconvex} and with the star-convex objective function in \Cref{thm:hess_convex}. The proofs of \Cref{thm:hess_nonconvex,thm:hess_convex} are available in \Cref{proof:hess_nonconvex,proof:hess_convex}, respectively.
\begin{theorem}\label{thm:hess_nonconvex}
  Let \Cref{eq:L,eq:variance,eq:norm,eq:H} hold, let $x_0 \in \dom R$ and $m_0 = g(x_0,\xi_0)$, and let $\beta = 0$ and $\gamma = 1/\alpha$. Then the iterations of \Cref{alg:hess} satisfy the following inequality:
  \begin{equation}
    \E{\min_{k=1,\ldots,K}\norm{\nabla f(x_k) + \subgrad R_k}_*}
    \leq
    \frac{\Delta_0}{\eta K}
    +\frac{7L\eta}{2}
    +\frac{H\eta^2}{\alpha^2}
    +\frac{2\rho\sigma}{\alpha K}
    +2\sqrt{\alpha}\rho\sigma,
  \end{equation}
  where $\subgrad R_k \in \partial R(x_k)$, $\Delta_0 = F(x_0) - \inf_x F(x)$.
\end{theorem}
\begin{theorem}\label{thm:hess_convex}
  Let \Cref{eq:L,eq:variance,eq:norm,eq:star_cvx,eq:H} hold, and let $x_0 \in \dom R$ and $m_0 = g(x_0,\xi_0)$. Let the parameters $\eta$ and $\beta$ satisfy \cref{eq:eta_beta}, and let $\gamma = 1/\alpha$. Then the output of \Cref{alg:hess} satisfies
  \begin{equation}
    \E{F(x_K) - F(x^*)}
    \leq
    (1-\beta)^K (F(x_0) - F(x^*))
    +2\eta\rho\sigma\left(\frac{1}{\alpha} + \frac{\sqrt{\alpha}}{\beta}\right)
    +\frac{4L\eta^2}{\beta} +\frac{4H\eta^3}{\alpha^2\beta}.
  \end{equation}
\end{theorem}
Using \Cref{thm:hess_nonconvex,thm:hess_convex}, it is not hard to obtain the explicit iteration complexities \Cref{alg:hess} in \Cref{cor:tr_decay,cor:str_decay} for non-convex and star-convex functions, respectively. The proofs are omitted due to their simplicity.
\begin{corollary}\label{cor:hess_nonconvex}
  To reach the precision $\E{\min_{k=1\ldots K} \norm{\nabla f(x_k) + \subgrad R_k}_*} \leq \epsilon$ by \Cref{alg:hess} under the conditions of \Cref{thm:hess_nonconvex}, it is sufficient to choose the parameters as follows:
  \begin{align}
    \eta &= \cO\left(\min\left\{
        \frac{\epsilon}{L},
        \frac{\epsilon^{1/2}}{H^{1/2}},
        \frac{\epsilon^{5/2}}{\rho^2\sigma^2H^{1/2}}
    \right\}\right),
    \quad
    \alpha = \cO\left(\min\left\{1, \frac{\epsilon^2}{\rho^2\sigma^2}\right\}\right),
    \quad
    \beta = 0,
    \\
    K &= \tilde{\cO}\left(\max\left\{
        \frac{\rho\sigma}{\epsilon},
        \frac{\rho^3\sigma^3}{\epsilon^3},
        \frac{L\Delta_0}{\epsilon^2},
        \frac{H^{1/2}\Delta_0}{\epsilon^{3/2}},
        \frac{H^{1/2}\Delta_0\rho^2\sigma^2}{\epsilon^{7/2}}
    \right\}\right).
  \end{align}
\end{corollary}
\begin{corollary}\label{cor:hess_convex}
  Under the conditions of \Cref{thm:hess_convex}, let $\norm{x_0} \leq \norm{x^*} = D$. To reach the precision $\E{F(x_K) - F(x^*)} \leq \epsilon$ by \Cref{alg:hess}, it is sufficient to choose the parameters as follows:
  \begin{align}
    \beta &= \cO\left(\min\left\{
        1,
        \frac{\epsilon}{LD^2},
        \frac{\alpha\epsilon}{D\rho\sigma},
        \frac{\alpha\epsilon^{1/2}}{H^{1/2}D^{3/2}}
    \right\}\right),
    \qquad\quad
    \alpha = \cO\left(\min\left\{1, \frac{\epsilon^2}{D^2\rho^2\sigma^2}\right\}\right),
    \\
    K &= \tilde{\cO}\left(\max\left\{
        1,
        \frac{D\rho\sigma}{\epsilon},
        \frac{D^3\rho^3\sigma^3}{\epsilon^3},
        \frac{LD^2}{\epsilon},
        \frac{H^{1/2}D^{3/2}}{\epsilon^{1/2}},
        \frac{H^{1/2}D^{7/2}\rho^2\sigma^2}{\epsilon^{5/2}}
    \right\}\right),
    \quad
    \eta = \beta D.
  \end{align}
\end{corollary}

\Cref{cor:hess_nonconvex} establishes an improved $1/\epsilon^{3.5}$ iteration complexity compared to the $1/\epsilon^4$ complexity of \Cref{alg} in \Cref{cor:str_nonconvex}, matching the results of \citet{cutkosky2020momentum,sun2023momentum} for normalized SGD and signSGD for non-convex second-order smooth functions.

Unfortunately, \Cref{alg:hess} cannot achieve a major improvement over \Cref{alg:decay} in the case of star-convex functions, since both have the term $(\rho\sigma D/\epsilon)^3$ in their complexities in \Cref{cor:str_decay,cor:hess_convex}, respectively. However, the practical success of Muon and the discussion in \cref{sec:example} may indicate that the ``geometric'' terms, which depend on the Lipschitz constants $L$ and $H$, play a more important role compared to the purely ``stochastic'' terms, which depend only on the variance $\sigma$. Hence, the improvement in the "mixed" term $\sqrt{\rho^4 \sigma^4HD^{7}/\epsilon^{5}}$ in \Cref{cor:hess_convex} over the corresponding term $\rho^2\sigma^2LD^3/\epsilon^3$ in \Cref{cor:str_decay} may be significant. Overall, it is an important open theoretical and practical question whether the extrapolation step~\eqref{eq:ox_hess} can improve the convergence of Muon.

\newpage

\bibliographystyle{apalike}
\bibliography{references}

\begin{thebibliography}{}

\bibitem[Achiam et~al., 2023]{achiam2023gpt}
Achiam, J., Adler, S., Agarwal, S., Ahmad, L., Akkaya, I., Aleman, F.~L., Almeida, D., Altenschmidt, J., Altman, S., Anadkat, S., et~al. (2023).
\newblock Gpt-4 technical report.
\newblock {\em arXiv preprint arXiv:2303.08774}.

\bibitem[Anil et~al., 2023]{team2023gemini}
Anil, R., Borgeaud, S., Alayrac, J.-B., Yu, J., Soricut, R., Schalkwyk, J., Dai, A.~M., Hauth, A., Millican, K., et~al. (2023).
\newblock Gemini: a family of highly capable multimodal models.
\newblock {\em arXiv preprint arXiv:2312.11805}.

\bibitem[Arjevani et~al., 2023]{arjevani2023lower}
Arjevani, Y., Carmon, Y., Duchi, J.~C., Foster, D.~J., Srebro, N., and Woodworth, B. (2023).
\newblock Lower bounds for non-convex stochastic optimization.
\newblock {\em Mathematical Programming}, 199(1):165--214.

\bibitem[Arjovsky et~al., 2017]{arjovsky2017wasserstein}
Arjovsky, M., Chintala, S., and Bottou, L. (2017).
\newblock Wasserstein generative adversarial networks.
\newblock In {\em International conference on machine learning}, pages 214--223. PMLR.

\bibitem[Bernstein and Newhouse, 2024]{bernstein2024old}
Bernstein, J. and Newhouse, L. (2024).
\newblock Old optimizer, new norm: An anthology.
\newblock {\em arXiv preprint arXiv:2409.20325}.

\bibitem[Bernstein et~al., 2020]{bernstein2020learning}
Bernstein, J., Zhao, J., Meister, M., Liu, M.-Y., Anandkumar, A., and Yue, Y. (2020).
\newblock Learning compositional functions via multiplicative weight updates.
\newblock {\em Advances in neural information processing systems}, 33:13319--13330.

\bibitem[Carlson et~al., 2015a]{carlson2015stochastic}
Carlson, D., Cevher, V., and Carin, L. (2015a).
\newblock Stochastic spectral descent for restricted boltzmann machines.
\newblock In {\em Artificial Intelligence and Statistics}, pages 111--119. PMLR.

\bibitem[Carlson et~al., 2015b]{carlson2015preconditioned}
Carlson, D.~E., Collins, E., Hsieh, Y.-P., Carin, L., and Cevher, V. (2015b).
\newblock Preconditioned spectral descent for deep learning.
\newblock {\em Advances in neural information processing systems}, 28.

\bibitem[Carmon et~al., 2020]{carmon2020lower}
Carmon, Y., Duchi, J.~C., Hinder, O., and Sidford, A. (2020).
\newblock Lower bounds for finding stationary points i.
\newblock {\em Mathematical Programming}, 184(1):71--120.

\bibitem[Conn et~al., 2000]{conn2000trust}
Conn, A.~R., Gould, N.~I., and Toint, P.~L. (2000).
\newblock {\em Trust region methods}.
\newblock SIAM.

\bibitem[Cutkosky and Mehta, 2020]{cutkosky2020momentum}
Cutkosky, A. and Mehta, H. (2020).
\newblock Momentum improves normalized sgd.
\newblock In {\em International conference on machine learning}, pages 2260--2268. PMLR.

\bibitem[Duchi et~al., 2011]{duchi2011adaptive}
Duchi, J., Hazan, E., and Singer, Y. (2011).
\newblock Adaptive subgradient methods for online learning and stochastic optimization.
\newblock {\em Journal of machine learning research}, 12(7).

\bibitem[Elsayed et~al., 2024]{elsayed2024weight}
Elsayed, M., Lan, Q., Lyle, C., and Mahmood, A.~R. (2024).
\newblock Weight clipping for deep continual and reinforcement learning.
\newblock {\em arXiv preprint arXiv:2407.01704}.

\bibitem[Ghadimi and Lan, 2013]{ghadimi2013stochastic}
Ghadimi, S. and Lan, G. (2013).
\newblock Stochastic first-and zeroth-order methods for nonconvex stochastic programming.
\newblock {\em SIAM journal on optimization}, 23(4):2341--2368.

\bibitem[Ghadimi and Lan, 2016]{ghadimi2016accelerated}
Ghadimi, S. and Lan, G. (2016).
\newblock Accelerated gradient methods for nonconvex nonlinear and stochastic programming.
\newblock {\em Mathematical Programming}, 156(1):59--99.

\bibitem[Gorbunov et~al., 2020]{gorbunov2020linearly}
Gorbunov, E., Kovalev, D., Makarenko, D., and Richt{\'a}rik, P. (2020).
\newblock Linearly converging error compensated sgd.
\newblock {\em Advances in Neural Information Processing Systems}, 33:20889--20900.

\bibitem[Gower et~al., 2019]{gower2019sgd}
Gower, R.~M., Loizou, N., Qian, X., Sailanbayev, A., Shulgin, E., and Richt{\'a}rik, P. (2019).
\newblock Sgd: General analysis and improved rates.
\newblock In {\em International conference on machine learning}, pages 5200--5209. PMLR.

\bibitem[Grattafiori et~al., 2024]{grattafiori2024llama}
Grattafiori, A., Dubey, A., Jauhri, A., Pandey, A., Kadian, A., Al-Dahle, A., Letman, A., Mathur, A., Schelten, A., Vaughan, A., et~al. (2024).
\newblock The llama 3 herd of models.
\newblock {\em arXiv preprint arXiv:2407.21783}.

\bibitem[Grimmer, 2019]{grimmer2019convergence}
Grimmer, B. (2019).
\newblock Convergence rates for deterministic and stochastic subgradient methods without lipschitz continuity.
\newblock {\em SIAM Journal on Optimization}, 29(2):1350--1365.

\bibitem[Gupta et~al., 2018]{gupta2018shampoo}
Gupta, V., Koren, T., and Singer, Y. (2018).
\newblock Shampoo: Preconditioned stochastic tensor optimization.
\newblock In {\em International Conference on Machine Learning}, pages 1842--1850. PMLR.

\bibitem[Guzm{\'a}n and Nemirovski, 2015]{guzman2015lower}
Guzm{\'a}n, C. and Nemirovski, A. (2015).
\newblock On lower complexity bounds for large-scale smooth convex optimization.
\newblock {\em Journal of Complexity}, 31(1):1--14.

\bibitem[Horv{\'a}th et~al., 2023]{horvath2023stochastic}
Horv{\'a}th, S., Kovalev, D., Mishchenko, K., Richt{\'a}rik, P., and Stich, S. (2023).
\newblock Stochastic distributed learning with gradient quantization and double-variance reduction.
\newblock {\em Optimization Methods and Software}, 38(1):91--106.

\bibitem[Hunter and Lange, 2004]{hunter2004tutorial}
Hunter, D.~R. and Lange, K. (2004).
\newblock A tutorial on mm algorithms.
\newblock {\em The American Statistician}, 58(1):30--37.

\bibitem[Jiang et~al., 2023]{jiang2023beyond}
Jiang, Y., He, C., Zhang, C., Ge, D., Jiang, B., and Ye, Y. (2023).
\newblock Beyond nonconvexity: A universal trust-region method with new analyses.
\newblock {\em arXiv e-prints}, pages arXiv--2311.

\bibitem[Jordan et~al., 2024]{jordan2024muon}
Jordan, K., Jin, Y., Boza, V., You, J., Cesista, F., Newhouse, L., and Bernstein, J. (2024).
\newblock Muon: An optimizer for hidden layers in neural networks.

\bibitem[Kingma and Ba, 2014]{kingma2014adam}
Kingma, D.~P. and Ba, J. (2014).
\newblock Adam: A method for stochastic optimization.
\newblock {\em arXiv preprint arXiv:1412.6980}.

\bibitem[Kleinberg et~al., 2018]{kleinberg2018alternative}
Kleinberg, B., Li, Y., and Yuan, Y. (2018).
\newblock An alternative view: When does sgd escape local minima?
\newblock In {\em International conference on machine learning}, pages 2698--2707. PMLR.

\bibitem[LeCun et~al., 2015]{lecun2015deep}
LeCun, Y., Bengio, Y., and Hinton, G. (2015).
\newblock Deep learning.
\newblock {\em nature}, 521(7553):436--444.

\bibitem[Liu et~al., 2024]{liu2024deepseek}
Liu, A., Feng, B., Xue, B., Wang, B., Wu, B., Lu, C., Zhao, C., Deng, C., Zhang, C., Ruan, C., et~al. (2024).
\newblock Deepseek-v3 technical report.
\newblock {\em arXiv preprint arXiv:2412.19437}.

\bibitem[Liu et~al., 2025]{liu2025muon}
Liu, J., Su, J., Yao, X., Jiang, Z., Lai, G., Du, Y., Qin, Y., Xu, W., Lu, E., Yan, J., et~al. (2025).
\newblock Muon is scalable for llm training.
\newblock {\em arXiv preprint arXiv:2502.16982}.

\bibitem[Loshchilov and Hutter, 2017]{loshchilov2017decoupled}
Loshchilov, I. and Hutter, F. (2017).
\newblock Decoupled weight decay regularization.
\newblock {\em arXiv preprint arXiv:1711.05101}.

\bibitem[Mokhtari et~al., 2020]{mokhtari2020stochastic}
Mokhtari, A., Hassani, H., and Karbasi, A. (2020).
\newblock Stochastic conditional gradient methods: From convex minimization to submodular maximization.
\newblock {\em Journal of machine learning research}, 21(105):1--49.

\bibitem[Nesterov, 2013]{nesterov2013gradient}
Nesterov, Y. (2013).
\newblock Gradient methods for minimizing composite functions.
\newblock {\em Mathematical programming}, 140(1):125--161.

\bibitem[Nesterov et~al., 2018]{nesterov2018lectures}
Nesterov, Y. et~al. (2018).
\newblock {\em Lectures on convex optimization}, volume 137.
\newblock Springer.

\bibitem[Nesterov and Polyak, 2006]{nesterov2006cubic}
Nesterov, Y. and Polyak, B.~T. (2006).
\newblock Cubic regularization of newton method and its global performance.
\newblock {\em Mathematical programming}, 108(1):177--205.

\bibitem[Nesterov, 1984]{nesterov1984minimization}
Nesterov, Y.~E. (1984).
\newblock Minimization methods for nonsmooth convex and quasiconvex functions.
\newblock {\em Matekon}, 29(3):519--531.

\bibitem[Pethick et~al., 2025]{pethick2025training}
Pethick, T., Xie, W., Antonakopoulos, K., Zhu, Z., Silveti-Falls, A., and Cevher, V. (2025).
\newblock Training deep learning models with norm-constrained lmos.
\newblock {\em arXiv preprint arXiv:2502.07529}.

\bibitem[Reddi et~al., 2019]{reddi2019convergence}
Reddi, S.~J., Kale, S., and Kumar, S. (2019).
\newblock On the convergence of adam and beyond.
\newblock {\em arXiv preprint arXiv:1904.09237}.

\bibitem[Rockafellar, 1997]{rockafellar1997convex}
Rockafellar, R.~T. (1997).
\newblock {\em Convex analysis}, volume~28.
\newblock Princeton university press.

\bibitem[Shor, 2012]{shor2012minimization}
Shor, N.~Z. (2012).
\newblock {\em Minimization methods for non-differentiable functions}, volume~3.
\newblock Springer Science \& Business Media.

\bibitem[Sun et~al., 2023]{sun2023momentum}
Sun, T., Wang, Q., Li, D., and Wang, B. (2023).
\newblock Momentum ensures convergence of signsgd under weaker assumptions.
\newblock In {\em International Conference on Machine Learning}, pages 33077--33099. PMLR.

\bibitem[Tieleman, 2012]{tieleman2012lecture}
Tieleman, T. (2012).
\newblock Lecture 6.5-rmsprop: Divide the gradient by a running average of its recent magnitude.
\newblock {\em COURSERA: Neural networks for machine learning}, 4(2):26.

\bibitem[Tuddenham et~al., 2022]{tuddenham2022orthogonalising}
Tuddenham, M., Pr{\"u}gel-Bennett, A., and Hare, J. (2022).
\newblock Orthogonalising gradients to speed up neural network optimisation.
\newblock {\em arXiv preprint arXiv:2202.07052}.

\bibitem[Zhou et~al., 2019]{zhou2019sgd}
Zhou, Y., Yang, J., Zhang, H., Liang, Y., and Tarokh, V. (2019).
\newblock Sgd converges to global minimum in deep learning via star-convex path.
\newblock {\em arXiv preprint arXiv:1901.00451}.

\bibitem[Zhuang et~al., 2022]{zhuang2022understanding}
Zhuang, Z., Liu, M., Cutkosky, A., and Orabona, F. (2022).
\newblock Understanding adamw through proximal methods and scale-freeness.
\newblock {\em arXiv preprint arXiv:2202.00089}.

\end{thebibliography}

\newpage
\appendix

\section{Additional Discussions}\label{sec:discussion}

\subsection{Motivition for the Non-Euclidean Smoothness Assumptions}\label{sec:example}

In this section, we provide a short discussion on the benefits of using the matrix spectral norm, $\norm{\cdot} = \norms{\cdot}$, in the training of neural networks. In particular, we consider the following simple example of \cref{eq:mat}, which motivates the smoothness \Cref{eq:L,eq:H}:
\begin{equation}\label[problem]{eq:example}
  \min_{\mX \in \sM} \left[F(\mX) = \frac{1}{N}\sum_{i=1}^N \cL_i(\mX a_i)\right],
\end{equation}
where $a_i \in \R^n$ are some vectors, and $\cL_i(\cdot)\colon \R^m \to \R$ are loss functions. Our example can be seen as a generalization of the example in Proposition~6 by \citet{bernstein2024old} and is also inspired by some earlier works of \citet{carlson2015preconditioned,carlson2015stochastic}. This example holds interest in the context of deep learning due to the matrix-vector multiplication term $\mX a_i$, because similar structures appear in hidden layers of neural networks.

We show that \Cref{eq:L,eq:H} are implied by the standard first- and second-order smoothness assumptions of functions $\cL_i(\cdot)$ with respect to the standard Euclidean norm $\norme{\cdot}$ in the following \Cref{lem:example_L,lem:example_H}, respectively. The proofs are available in \Cref{proof:example_L,proof:example_H}. The main observation here is that the smoothness constants $L$ and $H$ depend on the Euclidean norm $\norme{\mX a_i}$, which can be naturally upper-bounded using the spectral norm as $\norme{\mX a_i} \leq \norms{\mX}\norme{a_i}$. Note that a similar upper bound can be obtained using the standard Euclidean, aka Frobenius, norm of matrix $\mX$, which, however, would be much less tight. Also, note that besides the matrix spectral norm $\norms{\cdot}$, it is also possible to obtain similar upper bounds using more general operator norms. A more detailed study of this question is provided by \citet{pethick2025training}.

\begin{lemma}\label{lem:example_L}
  Let each function $\cL_i(\cdot)$ in \cref{eq:example} be continuously differentiable and have a $\lambda$-Lipschitz gradient with respect to the standard Euclidean norm $\norme{\cdot}$. Then function $F(\cdot)$ satisfies \Cref{eq:L} with the matrix spectral norm, $\norm{\cdot} = \norms{\cdot}$, and the gradient Lipschitz constant
  \begin{equation}\label{eq:example_L}
    L = \lambda \cdot \frac{1}{N}\sum_{i=1}^N \sqne{a_i}.
  \end{equation}
\end{lemma}
\begin{lemma}\label{lem:example_H}
  Let each function $\cL_i(\cdot)$ in \cref{eq:example} be twice continuously differentiable and have a $\lambda$-Lipschitz Hessian with respect to the standard Euclidean norm $\norme{\cdot}$. Then function $F(\cdot)$ satisfies \Cref{eq:H} with the matrix spectral norm, $\norm{\cdot} = \norms{\cdot}$, and the Hessian Lipschitz constant
  \begin{equation}\label{eq:example_H}
    H = \lambda \cdot \frac{1}{N}\sum_{i=1}^N \norme{a_i}^3.
  \end{equation}
\end{lemma}

\subsection[Comparison with Pethick et al. (2025)]{Comparison with \cite{pethick2025training}}\label{sec:FW}

In this section, we provide a brief comparison of our results with the concurrent work of \citet{pethick2025training}. We discovered this work only after the first version of our paper appeared online. In contrast to our trust-region approach, \citet{pethick2025training} build their analysis on the stochastic conditional gradient method of \citet{mokhtari2020stochastic}. Both these approaches are closely related, and we leave further investigation of this connection for future work. Besides, they provide analysis for arbitrary non-Euclidean norms, just as we do. However, their theoretical results have the following disadvantages:
\begin{itemize}[topsep=0pt,parsep=0pt,partopsep=0pt]
  \item[\bf (i)] The analysis of \cite{pethick2025training} is limited to the case where the regularizer $R(\cdot)$ is the indicator function of some closed and convex set. In contrast, our analysis works with an arbitrary proper, closed, and convex regularizer.
  \item[\bf (ii)] \citet{pethick2025training} obtain the same $1/\epsilon^4$ complexity for Muon, both with and without weight decay. In particular, they do not obtain the improved $1/\epsilon^3$ complexity under the star-convexity assumption. Besides, their theoretical results provide a limited explanation for the success of Muon with weight decay in the training of large-scale language models \citep{liu2025muon}, and for the success of Muon without weight decay in the training of small-scale language models \citep{jordan2024muon}. In contrast, we discuss both cases in \Cref{sec:decay_convex,sec:D}.
  \item[\bf (iii)] In the case of non-convex functions, \citet{pethick2025training} establish only the complexity $1/\epsilon^4$, which is standard for non-convex functions. In contrast, we obtain the improved $1/\epsilon^{3.5}$ complexity by utilizing an additional extrapolation step and the second-order smoothness assumption.
\end{itemize}

\subsection{Complexities for Non-Convex and Star-Convex Functions}\label{sec:cmp}

In this paper, we often compare the complexities of finding an $\epsilon$-accurate solution to \cref{eq:main} with non-convex and star-convex objective functions $f(x)$. It is important to highlight that we use the approximate first-order stationarity $\norm{\nabla f(x_k) + \subgrad R_k}_* \leq \epsilon$ for non-convex functions, and the approximate functional suboptimality $F(x_k) - F(x^*) \leq \epsilon$.
For instance, in \Cref{sec:str_nonconvex}, comparing the $1/\epsilon^3$ complexity result in \Cref{cor:str_decay} for star-convex functions with the $1/\epsilon^{3.5}$ result of \citet{cutkosky2020momentum,sun2023momentum} for non-convex functions may seem incorrect.

However, by slightly modifying the proof of \Cref{thm:str_D}, we can obtain the same improved complexity $1/\epsilon^3$ for finding an $\epsilon$-stationary point under the star-convexity \Cref{eq:star_cvx} and bounded domain \Cref{eq:D}. This can be correctly compared with the results for non-convex functions, such as \Cref{cor:hess_nonconvex,cor:str_nonconvex}. On the other hand, under these assumptions, we can upper-bound the functional suboptimality as follows:
\begin{equation}
  F(x_k) - F(x^*) \leq D\norm{\nabla f(x_k) + \subgrad R_k}_*.
\end{equation}
Hence, the generalized $(\epsilon/D)$-stationarity $\norm{\nabla f(x_k) + \subgrad R_k}_*$ implies the $\epsilon$-approximate functional suboptimality as well. Note that, strictly speaking, these considerations are not valid for the case of zero regularization, $R\equiv 0$. We leave further investigation for future work.



\subsection{Proof of \Cref{lem:example_L}}\label{proof:example_L}

The inequality in \Cref{eq:L} is implied by the following inequality, which we further prove:
\begin{equation}
  \nabla F(\mX_1)[\mH] - \nabla F(\mX_2)[\mH] \leq L \norms{\mX_1 - \mX_2}\norms{\mH},
\end{equation}
where $\mX_1, \mH \in \sM$. We proceed as follows:
\begin{align*}
  \mind{3em}
  \nabla F(\mX_1)[\mH] - \nabla F(\mX_2)[\mH]
  \\&\aleq{uses the linearity of the first differential}
  \frac{1}{N}\sum_{i=1}^N
  \left(\nabla \cL_i(\mX_1 a_i)[\mH a_i] - \nabla \cL_i(\mX_2 a_i)[\mH a_i]\right)
  \\&\aleq{uses the $\lambda$-Lipschitzness of the gradients $\nabla \cL_i$ with respect to the standard Euclidean norm $\norme{\cdot}$}
  \frac{1}{N}\sum_{i=1}^N
  \lambda \norme{\mX_1 a_i - \mX_2 a_i}\norme{\mH a_i}
  \\&\aleq{uses the properties of the matrix spectral norm $\norms{\cdot}$}
  \frac{1}{N}\sum_{i=1}^N \lambda\norms{\mX_1 - \mX_2}\norms{\mH} \sqne{a_i}
  \\&\aeq{uses the definition of $L$ in \cref{eq:example_L}}
  L \norms{\mX_1 - \mX_2}\norms{\mH},
\end{align*}
where \annotate.\qed

\subsection{Proof of \Cref{lem:example_H}}\label{proof:example_H}

The inequality in \Cref{eq:H} is implied by the following inequality, which we further prove:
\begin{equation}
  \nabla^2 F(\mX_1)[\mX_1 - \mX_2,\mH] - \nabla^2 F(\mX_2)[\mX_1 - \mX_2,\mH] \leq H \sqns{\mX_1 - \mX_2}\norms{\mH},
\end{equation}
where $\mX_1,\mX_2, \mH \in \sM$. We proceed as follows:
\begin{align*}
  \mind{3em}
  \nabla^2 F(\mX_1)[\mX_1 - \mX_2,\mH] - \nabla^2 F(\mX_2)[\mX_1 - \mX_2,\mH]
  \\&\aleq{uses the multilinearity of the second differential}
  \frac{1}{N}\sum_{i=1}^N
  \left(\nabla^2 \cL_i(\mX_1 a_i)[(\mX_1 - \mX_2)a_i,\mH a_i] - \nabla^2 \cL_i(\mX_2 a_i)[(\mX_1 - \mX_2)a_i,\mH a_i]\right)
  \\&\aleq{uses the $\lambda$-Lipschitzness of the Hessians $\nabla^2 \cL_i$ with respect to the standard Euclidean norm $\norme{\cdot}$}
  \frac{1}{N}\sum_{i=1}^N
  \lambda \norme{\mX_1 a_i - \mX_2 a_i}\norme{(\mX_1 - \mX_2)a_i}\norme{\mH a_i}
  \\&\aleq{uses the properties of the matrix spectral norm $\norms{\cdot}$}
  \frac{1}{N}\sum_{i=1}^N \lambda\sqns{\mX_1 - \mX_2}\norms{\mH} \norme{a_i}^3
  \\&\aeq{uses the definition of $H$ in \cref{eq:example_H}}
  H \sqns{\mX_1 - \mX_2}\norms{\mH},
\end{align*}
where \annotate.\qed

\section{Muon vs Orthogonal-SGDM}\label{sec:osgdm}

As previously discussed in \Cref{sec:alg}, one can verify that in the case of the non-Euclidean spectral norm $\norm{\cdot} = \norms{\cdot}$ and zero regularizer $R \equiv 0$, \Cref{alg} exactly matches the Muon optimizer as it was originally described by \citet{jordan2024muon}. As mentioned in \Cref{sec:motivation}, there is another variant of the orthogonalized gradient method with momentum proposed by \citet{tuddenham2022orthogonalising}, namely, Orthogonal-SGDM. We provide pseudocode for both algorithms in \Cref{alg:muon}. While Muon can achieve impressive practical results \citep{jordan2024muon,liu2025muon}, Orthogonal-SGDM is unable to outperform Muon or well-tuned standard SGD with momentum \citep{tuddenham2022orthogonalising,jordan2024muon}. In this section, we provide a possible explanation for this phenomenon using our theoretical insights from \Cref{sec:str_nonconvex}.

\begin{algorithm}[H]
  \caption{Muon/Orthogonal-SGDM \citep{jordan2024muon,tuddenham2022orthogonalising}}
  \label{alg:muon}
  \begin{algorithmic}[1]
    \State {\bf input:} $\mX_0, \mM_0 \in \sM$
    \State {\bf parameters:} $\eta > 0$, $\alpha \in (0,1)$, $K \in \{1,2,\ldots\}$
    \For{$k=0,1,\ldots,K-1$}
    \State Compute $\mG_k = \nabla F(\mX_k)$
    \State Compute $\mX_{k+1}$ and $\mM_{k+1}$ as follows:
    \begin{alignat}{20}
      \label{eq:muon}
      \mX_{k+1} &= \mX_k - \eta \mO_{k+1},\;
      &
      \mM_{k+1} &= (1-\alpha)\mM_k + \alpha \mG_k,\;
      &
      \mO_{k+1} &= \orth(\mM_{k+1})\tag{Muon}
      \\
      \label{eq:osgdm}
      \mX_{k+1} &= \mX_k - \eta \mM_{k+1},\;
      &
      \mM_{k+1} &= (1-\alpha)\mM_k + \alpha \mO_{k+1},\;
      &
      \mO_{k+1} &= \orth(\mG_k)\tag{OSGDM}
    \end{alignat}
    \EndFor
    \State {\bf output:} $\mX_K \in \sM$
  \end{algorithmic}
\end{algorithm}

The main difference between the updates~\eqref{eq:muon} and~\eqref{eq:osgdm} is the order in which the momentum and orthogonalization are applied. Our theory, in particular \Cref{lem:descent,lem:momentum} in \Cref{proof:str_nonconvex}, suggests that the update rule~\eqref{eq:muon} is preferable:
\begin{itemize}[topsep=0pt,parsep=0pt,partopsep=0pt]
  \item[\bf (i)]
    First, by analyzing the proof of \Cref{lem:momentum}, one can conclude that the momentum accumulates a weighted sum of the terms $\zeta_k = \nabla f(x_k) - g(x_k,\xi_k)$. On the other hand, from \Cref{eq:variance}, it follows that the variance of $\zeta_k$ is bounded. What is even more important, this assumption implies that the random variables $\zeta_k$ are independent and centered, i.e., $\E{\zeta_k} = 0$, which means that the total variance of their weighted sum is reduced due to the so-called {\em batching effect}. Refer to the proof of \Cref{lem:momentum} in \Cref{proof:momentum} for details.
  \item[\bf (ii)]
    Second, recall our trust-region interpretation of the orthogonalized gradient descent. According to the discussion in \Cref{sec:alg}, it is natural to feed the momentum term $m_{k+1}$ to the trust-region optimization step~\eqref{eq:str} with the non-Euclidean spectral norm, which results in the <<orthogonalized momentum>> $\mO_{k+1}$ in \cref{eq:muon} rather than the <<momentum of orthogonals>> $\mM_{k+1}$ in \cref{eq:osgdm}. These considerations are supported by the proof of \Cref{lem:descent} in \Cref{proof:descent}.
\end{itemize}
In contrast, the considerations above are not applicable to the update rule~\eqref{eq:osgdm}. Indeed, there is no reason to believe that the terms $\mO_{k+1}$ in \cref{eq:osgdm} are unbiased in any sense or that their accumulated variance can be reduced. Additionally, it is unclear what the possible interpretation of the momentum term $\mM_{k+1}$, constructed from the orthogonalized parts $\mO_{k+1}$ in \cref{eq:osgdm}, could be.

\newpage

\section{Algorithms with Weight Clipping for Star-Convex Functions}\label{sec:D}

As mentioned in \Cref{sec:alg_decay}, it is possible to obtain improved convergence guarantees for the iterations~\eqref{eq:tr} under the star-convexity \Cref{eq:star_cvx}, provided that these iterations are bounded. Recall that the iterations~\eqref{eq:tr_decay} and \Cref{alg:decay} use weight decay to achieve this. An alternative approach is to simply assume the boundedness of $\dom R$, which implies the following inequality:
\begin{aequation}\label{eq:D}
  \norm{x-x'} \leq D
  \quad\text{for all}\;x,x' \in \dom R,
\end{aequation}
where $D > 0$ is the diameter of $\dom R$. In this section, we provide convergence guarantees for the iterations~\eqref{eq:tr} and \Cref{alg} for star-convex functions under the additional \Cref{eq:D}.

We start with the result in \Cref{thm:tr_D} for the deterministic iterations~\eqref{eq:tr}. The proof is available in \Cref{proof:tr_D}. Next, we obtain the convergence result for the stochastic \Cref{alg} in \Cref{thm:str_D}. The proof is available in \Cref{proof:str_D}. Finally, using \Cref{thm:tr_D,thm:str_D}, it is not hard to obtain the explicit iteration complexities for the iterations~\eqref{eq:tr} and \Cref{alg} in \Cref{cor:tr_D,cor:str_D}, respectively. The proofs are omitted due to their simplicity.
\begin{theorem}\label{thm:tr_D}
  Let \Cref{eq:L,eq:D,eq:star_cvx} hold, and let $x_0 \in \dom R$.
  Then the iterations~\eqref{eq:tr} satisfy the following inequality:
  \begin{equation}
    F(x_K) - F(x^*)
    \leq
    \left(\frac{D}{\eta+D}\right)^K (F(x_0) - F(x^*)) + \frac{3LD\eta}{2}.
  \end{equation}
\end{theorem}
\begin{theorem}\label{thm:str_D}
  Let \Cref{eq:L,eq:variance,eq:norm,eq:star_cvx,eq:D} hold, and let $x_0 \in \dom R$ and $m_0 = g(x_0,\xi_0)$. Then the iterations of \Cref{alg} satisfy the following inequality:
  \begin{equation}
    \E{F(x_K) - F(x^*)}
    \leq
    \left(\frac{D}{\eta + D}\right)^K(F(x_0) - F(x^*))
    +2\sqrt{\alpha}D\rho\sigma
    +\frac{2\eta\rho\sigma}{\alpha}
    +\frac{3LD\eta}{2}
    +\frac{2LD\eta}{\alpha}.
  \end{equation}
\end{theorem}
\begin{corollary}\label{cor:tr_D}
  Under the conditions of \Cref{thm:tr_D}, to reach the precision $F(x_K) - F(x^*) \leq \epsilon$ by the iterations~\eqref{eq:tr}, it is sufficient to choose the stepsize $\eta$ and the number of iterations $K$ as follows:
  \begin{equation}
    \eta = \cO\left(\frac{\epsilon}{LD}\right),\qquad
    K = \tilde{\cO}\left(\max\left\{1,\frac{LD^2}{\epsilon}\right\}\right).
  \end{equation}
\end{corollary}
\begin{corollary}\label{cor:str_D}
  Under the conditions of \Cref{thm:str_D}, to reach the precision $\E{F(x_K) - F(x^*)} \leq \epsilon$ by \Cref{alg}, it is sufficient to choose the parameters as follows:
  \begin{align}
    \eta &= \cO\left(\min\left\{
        \frac{\epsilon}{LD},
        \frac{\epsilon}{\rho\sigma},
        \frac{\epsilon^3}{D^2\rho^3\sigma^3},
        \frac{\epsilon^3}{LD^3\rho^2\sigma^2}
    \right\}\right)
    ,\qquad
    \alpha = \cO\left(\min\left\{1, \frac{\epsilon^2}{D^2\rho^2\sigma^2}\right\}\right),
    \\
    K &= \tilde{\cO}\left(\max\left\{
        1,
        \frac{LD^2}{\epsilon},
        \frac{D\rho\sigma}{\epsilon},
        \frac{D^3\rho^3\sigma^3}{\epsilon^3},
        \frac{LD^4\rho^2\sigma^2}{\epsilon^3}
    \right\}\right).
  \end{align}
\end{corollary}

The complexity of the iterations~\eqref{eq:tr} without weight decay in \Cref{cor:tr_D} matches the complexity of the iterations~\eqref{eq:D} with weight decay in \Cref{cor:tr_decay}. Similarly, the complexity of \Cref{alg} without weight decay in \Cref{cor:str_D} matches the result for \Cref{alg:decay} with weight decay in \Cref{cor:str_decay}.
It is important to highlight that \Cref{eq:D} obviously does not hold in the case of zero regularization $R\equiv 0$. To tackle this issue, one could choose the regularizer $R(\cdot)$ as the following indicator function:
\begin{equation}\label{eq:reg_clip}
  R(x) =
  \begin{cases}
    0 & \norm{x} \leq D\\
    +\infty & \norm{x} > D
  \end{cases},\quad
  \text{where}\; D \geq \norm{x^*}.
\end{equation}
In the case of the infinity norm $\norm{\cdot} = \infnorm{\cdot}$, \Cref{alg} with the regularizer~\eqref{eq:reg_clip} can be easily implemented and reduces to signSGD with momentum and weight clipping. Weight clipping is used in various machine learning applications \citep{elsayed2024weight,bernstein2020learning,arjovsky2017wasserstein} and can be seen as an alternative to weight decay used in \Cref{alg:decay}.

Unfortunately, in the case of the spectral norm $\norm{\cdot} = \norms{\cdot}$, \Cref{alg} with the regularizer~\eqref{eq:reg_clip}, such a clipping procedure is not easily implementable, which justifies the use of weight decay in Muon. On the other hand, \citet{liu2025muon} reported that in cases where language models are small-scale and/or the training time is not lengthy, the iterations of Muon do not grow excessively large. Hence, with a reasonable choice of the radius $D$, the weight clipping step in \Cref{alg} would be skipped. Thus, the fact that we obtain the same complexity results for \Cref{alg} with weight clipping and \Cref{alg:decay} weight decay may support the practical evidence that weight decay is not necessary for the training of small-scale language models \citep{liu2025muon,jordan2024muon}.

\newpage

\section{Proofs of Theorems}

\subsection{Proof of \Cref{thm:tr_nonconvex}}\label{proof:tr_nonconvex}

We start with the following \Cref{lem:trust_reg_prox}, which allows us to incorporate an arbitrary proper, closed, and convex regularizer $R(x)$ into the non-Euclidean trust-region optimization step. The proof is available in \Cref{proof:trust_reg_prox}.
\begin{lemma}\label{lem:trust_reg_prox}
  Let $z_+ \in \sX$ be defined as a solution to the following trust-region optimization problem:
  \begin{equation}\label[problem]{eq:trust_reg_prox}
    z_+ = \argmin_{x\in \sX} \Psi(x) + R(x)
    \quad\text{s.t.}\quad
    \norm{x - z} \leq \eta,
  \end{equation}
  where $\Psi(\cdot)\colon \sX \to \R$ is a differentiable function, $z \in \dom R$ is a center of the trust-region, and $\eta > 0$ is the trust-region radius. Then the following inequality holds:
  \begin{equation}
    R(z) + \<\nabla \Psi(z_+),z - z_+>
    \geq
    R(z_+) + \eta\norm{\nabla \Psi(z_+) + \subgrad R}_*,
  \end{equation}
  where $\subgrad R \in \partial R(z_+)$.
\end{lemma}

Now, we are ready to prove \Cref{thm:tr_nonconvex}. We can upper-bound $F(x_{k+1})$ as follows:
\begin{align*}
  F(x_{k+1})
  &\aeq{use the definition of $F(x)$ in \cref{eq:main}}
  f(x_{k+1}) + R(x_{k+1})
  \\&\aleq{use \Cref{eq:L}}
  f(x_k) + \<\nabla f(x_k),x_{k+1} - x_k> +\tfrac{1}{2}L\sqn{x_{k+1} - x_k} + R(x_{k+1})
  \\&\aleq{uses \Cref{lem:trust_reg_prox} with $\Psi(x) = \<\nabla f(x_k),x>$ and $(z,z_+)=(x_k,x_{k+1})$}
  f(x_k) + R(x_k) - \eta\norm{\nabla f(x_k) + \subgrad R_{k+1}}_* + \tfrac{1}{2}L\sqn{x_{k+1} - x_k}
  \\&\aeq{uses the triangle inequality}
  f(x_k) + R(x_k) - \eta\norm{\nabla f(x_{k+1}) + \subgrad R_{k+1}}_* + \eta\norm{\nabla f(x_{k+1}) - \nabla f(x_k)}_* \\&+ \tfrac{1}{2}L\sqn{x_{k+1} - x_k}
  \\&\aleq{use \Cref{eq:L}}
  f(x_k) + R(x_k) - \eta\norm{\nabla f(x_{k+1}) + \subgrad R_{k+1}}_* + L\eta\norm{x_{k+1}-x_k} + \tfrac{1}{2}L\sqn{x_{k+1} - x_k}
  \\&\aleq{uses the fact that $\norm{x_{k+1} - x_k} \leq \eta$, which is implied by \cref{eq:tr}}
  f(x_k) + R(x_k) - \eta\norm{\nabla f(x_{k+1}) + \subgrad R_{k+1}}_* + \tfrac{3}{2}L\eta^2
  \\&\aeq{use the definition of $F(x)$ in \cref{eq:main}}
  F(x_k) - \eta\norm{\nabla f(x_{k+1}) + \subgrad R_{k+1}}_* + \tfrac{3}{2}L\eta^2,
\end{align*}
where \annotate.\qed

\subsubsection{Proof of \Cref{lem:trust_reg_prox}}\label{proof:trust_reg_prox}
The objective function in \cref{eq:trust_reg_prox} is lower semi-continuous, because function $R(x)$ is proper, closed and convex, and the constrained set in this problem is obviously compact. Hence, there exists at least a single solution $z^+ \in \sX$ to the problem.

Let function $I_\eta(x)$ be the indicator function of the constraint set in \cref{eq:trust_reg_prox}, i.e.,
\begin{equation}
  I_\eta(x) =
  \begin{cases}
    0& \norm{x - z} \leq \eta\\
    +\infty & \norm{x - z} > \eta
  \end{cases},
\end{equation}
and let function $R_\eta(x)$ be defined as follows:
\begin{equation}\label{aux:5}
  R_\eta(x) = R(x) + I_\eta(x).
\end{equation}
It is easy to verify that function $R_\eta(x)$ is proper, closed, and convex, and \cref{eq:trust_reg_prox} is equivalent to the following:
\begin{equation}\label[problem]{eq:trust_reg_prox2}
  \min_{x \in \sX} \Psi(x) + R_\eta(x).
\end{equation}
Therefore, $z_+$ satisfies the following inequality for all $x \in \sX$:
\begin{equation}\label{aux:3}
  \Psi(x) + R_\eta(x) \geq \Psi(z_+) + R_\eta(z_+).
\end{equation}

Our next goal is to show that $-\nabla \Psi(z_+) \in \partial R_\eta(z_+)$, which we prove by contradiction. Suppose that the opposite holds, i.e., $-\nabla \Psi(z_+) \notin \partial R_\eta(z_+)$. This implies the existence of $x \in \sX$ and $\delta > 0$ such that
\begin{equation}\label{aux:1}
  R_\eta(x) - R_\eta(z_+) + \<\nabla \Psi(z_+), x - z_+> = -\delta.
\end{equation}
On the other hand, by the definition of the differentiability, we have
\begin{equation}
  \lim_{\tau \to 0} \frac{\Psi(z_+ + \tau(x - z_+)) - \Psi(z_+) - \tau\<\nabla \Psi(z_+),x - z_+>}{\tau} = 0.
\end{equation}
Hence, there exists $\tau \in (0,1)$ such that
\begin{equation}\label{aux:2}
  \Psi(z_+ + \tau(x - z_+)) - \Psi(z_+) - \tau\<\nabla \Psi(z_+),x - z_+> \leq \tfrac{1}{2}\delta\tau.
\end{equation}
Hence, we obtain the following:
\begin{align*}
  -\delta
  &\aeq{uses \cref{aux:1}}
  R_\eta(x) - R_\eta(z_+) + \<\nabla \Psi(z_+), x - z_+>
  \\&\ageq{uses the convexity of function $R_\eta(x)$}
  \frac{1}{\tau}\big(R_\eta(z_+ + \tau(x - z_+)) - R_\eta(z_+) + \tau\<\nabla \Psi(z_+), x - z_+>\big)
  \\&\ageq{uses \cref{aux:3}}
  -\frac{1}{\tau}\big(\Psi(z_+ + \tau(x - z_+)) - \Psi(z_+) - \tau\<\nabla \Psi(z_+), x - z_+>\big)
  \\&\ageq{uses \cref{aux:2}}
  -\tfrac{1}{2}\delta,
\end{align*}
where \annotate. This implies the contradiction $\delta \leq 0$. Thus, we have shown that $-\nabla \Psi(z_+) \in \partial R_\eta(z_+)$.

Our next goal is to show that $\ri \dom R \cap \ri \dom I_\eta \neq \varnothing$. In the case where $z \in \ri \dom R$, this statement is obviously true because $z \in \ri \dom I_\eta$. Hence, it remains to consider the case $z \notin \ri \dom R$. By Theorem~6.3 of \citet{rockafellar1997convex}, we have $\closure(\ri \dom R) = \closure \dom R \supset \dom R$. Hence, there exists a sequence of vectors $\{x_k\}_{k=0}^{\infty}\subset \ri \dom R$ such that $\lim_{k \to \infty} \norm{x_k - z} = 0$. Since $I_\eta(x)$ is the indicator of the ball with the center at $z$, we can find $x_k \in \ri \dom I_\eta$. Thus, we have $x_k \in \ri \dom I_\eta \cap \ri\dom R \neq \varnothing$.

Using the definition of $R_\eta(x)$ in \cref{aux:5}, the fact that $\ri \dom R \cap \ri \dom I_\eta \neq \varnothing$, and Theorem~23.8 of \citet{rockafellar1997convex}, we have $\partial R_\eta(z_+) = \partial R(z_+) + \partial I_\eta(z_+)$. Hence, using the fact that $-\nabla \Psi(z_+) \in \partial R_\eta(z_+)$, we have
\begin{equation}
  \label{aux:4}
  \nabla \Psi(z_+) + \subgrad R + \subgrad I_\eta = 0,
\end{equation}
where $\subgrad R \in \partial R(z_+)$, and $\subgrad I_\eta \in \partial I_\eta(z_+)$. Therefore, we obtain the following:
\begin{align*}
  R(z) + \<\nabla \Psi(z_+),z - z_+>
  &\ageq{uses the convexity of function $R(z)$}
  R(z_+) + \<\nabla \Psi(z_+) + \subgrad R,z - z_+>
  \\&\aeq{use \cref{aux:4}}
  R(z_+) - \<\subgrad I_\eta,z - z_+>
  \\&=
  R(z_+) + \<\subgrad I_\eta,z_+> - \<\subgrad I_\eta,z>
  \\&\aeq{uses the Fenchel-Young equality}
  R(z_+) + I_\eta(z_+) + I_\eta^*(\subgrad I_\eta) - \<\subgrad I_\eta,z>
  \\&\aeq{uses the definition of $I_\eta(x)$ and standard calculations}
  R(z_+) + \eta\norm{\subgrad I_\eta}_*
  \\&\aeq{use \cref{aux:4}}
  R(z_+) + \eta\norm{\nabla \Psi(z_+) + \subgrad R}_*,
\end{align*}
where \annotate.\qed

\newpage

\subsection{Proof of \Cref{thm:str_nonconvex}}\label{proof:str_nonconvex}

We start with the following \Cref{lem:descent}, which describes the update rule~\eqref{eq:str}. It can be seen as an adaptation of the famous {\em descent lemma} in the context of trust-region optimization. The proof is, in many ways, based on \Cref{lem:trust_reg_prox} and is available in \Cref{proof:descent}.
\begin{lemma}\label{lem:descent}
  Let \Cref{eq:L} hold, and let $x_0 \in \dom R$. Then the iterations of \Cref{alg} satisfy the following inequality:
  \begin{equation}\label{eq:descent}
    F(x_{k+1})
    \leq
    F(x_k)
    -\eta\norm{\nabla f(x_{k+1}) + \subgrad R_{k+1}}_*
    +2\eta\norm{\nabla f(x_{k+1}) - m_{k+1}}_*
    + \tfrac{3}{2}L\eta^2,
  \end{equation}
  where $\subgrad R_{k+1} \in \partial R(x_{k+1})$.
\end{lemma}

The next \Cref{lem:momentum} describes the dynamics of the momentum term, which is updated according to \cref{eq:momentum}. In particular, it upper-bounds the expected distance between the momentum $m_k$ and the true gradient $\nabla f(x_k)$. The proof can be seen as an adaptation of the proof by \citet{cutkosky2020momentum} to the non-Euclidean norm setting and is available in \Cref{proof:momentum}.
\begin{lemma}\label{lem:momentum}
  Let \Cref{eq:L,eq:variance,eq:norm} hold, and let $x_0 \in \dom R$ and $m_0 = g(x_0,\xi_0)$. Then the iterations of \Cref{alg} satisfy the following inequality for $k \geq 0$:
  \begin{equation}
    \E{\norm{m_{k+1} - \nabla f(x_{k})}_*} \leq  (1-\alpha)^{k+1}\rho\sigma + \sqrt{\alpha}\rho\sigma + \frac{L\eta}{\alpha}.
  \end{equation}
\end{lemma}

Now, we are ready to prove \Cref{thm:str_nonconvex}. Using \Cref{lem:descent}, we obtain the following inequality:
\begin{align*}
  \min_{k=1,\ldots,K}\norm{\nabla f(x_k) + \subgrad R_k}_*
  &\leq
  \frac{F(x_0) - \inf_x F(x)}{\eta K} + \frac{3L\eta}{2} + \frac{2}{K}\sum_{k=1}^{K}\norm{\nabla f(x_k) - m_k}_*
  \\&\aleq{uses the triangle inequality, \Cref{eq:L}, and \cref{eq:str}}
  \frac{F(x_0) - \inf_x F(x)}{\eta K} + \frac{7L\eta}{2} + \frac{2}{K}\sum_{k=0}^{K-1}\norm{\nabla f(x_k) - m_{k+1}}_*,
\end{align*}
where \annotate.
Using \Cref{lem:momentum}, we obtain
\begin{align*}
  \E{\min_{k=1,\ldots,K}\norm{\nabla f(x_k) + \subgrad R_k}_*}
  \leq
  \frac{F(x_0) - \inf_x F(x)}{\eta K}
  +\frac{7L\eta}{2}
  +\frac{2 L\eta}{\alpha}
  +\frac{2\rho\sigma}{\alpha K}
  +2\sqrt{\alpha}\rho\sigma.
\end{align*}
\qed

\subsubsection{Proof of \Cref{lem:descent}}\label{proof:descent}
We can upper-bound $F(x_{k+1})$ as follows:
\begin{align*}
  F(x_{k+1})
  &\aeq{use the definition of function $F(x)$ in \cref{eq:main}}
  f(x_{k+1}) + R(x_{k+1})
  \\&\aleq{use \Cref{eq:L}}
  f(x_k) + \<\nabla f(x_k), x_{k+1} - x_k> + \tfrac{1}{2}L\sqn{x_{k+1} - x_k} + R(x_{k+1})
  \\&=
  f(x_k)+ \tfrac{1}{2}L\sqn{x_{k+1} - x_k} + R(x_{k+1})
  \\&
  + \<m_{k+1} + \nabla f(x_{k+1}) - m_{k+1} + \nabla f(x_k) - \nabla f(x_{k+1}), x_{k+1} - x_k>
  \\&\aleq{uses the definition of the dual norm}
  f(x_k)+ \tfrac{1}{2}L\sqn{x_{k+1} - x_k} + R(x_{k+1})
  +\<m_{k+1}, x_{k+1} - x_k>
  \\&
  +\norm{x_{k+1} - x_k}\norm{\nabla f(x_{k+1}) - m_{k+1}}_*
  +\norm{x_{k+1} - x_k}\norm{\nabla f(x_k) - \nabla f(x_{k+1})}_*
  \\&\aleq{use \Cref{eq:L}}
  f(x_k)+ \tfrac{3}{2}L\sqn{x_{k+1} - x_k}
  +\norm{x_{k+1} - x_k}\norm{\nabla f(x_{k+1}) - m_{k+1}}_*
  \\&
  + R(x_{k+1})+\<m_{k+1}, x_{k+1} - x_k>
  \\&\aleq{uses the fact that $\norm{x_{k+1} - x_k} \leq \eta$, which is implied by \cref{eq:str}}
  f(x_k)+ \tfrac{3}{2}L\eta^2
  +\eta\norm{\nabla f(x_{k+1}) - m_{k+1}}_*
  + R(x_{k+1})+\<m_{k+1}, x_{k+1} - x_k>
  \\&\aleq{uses \cref{eq:str} and \Cref{lem:trust_reg_prox} with $\Psi(x) = \<m_{k+1},x>$ and $(z_+,z) = (x_{k+1},x_k)$}
  f(x_k)+ \tfrac{3}{2}L\eta^2
  +\eta\norm{\nabla f(x_{k+1}) - m_{k+1}}_*
  + R(x_k) - \eta\norm{m_{k+1} + \subgrad R_{k+1}}_*
  \\&\aeq{use the definition of function $F(x)$ in \cref{eq:main}}
  F(x_k)+ \tfrac{3}{2}L\eta^2
  +\eta\norm{\nabla f(x_{k+1}) - m_{k+1}}_*
  - \eta\norm{m_{k+1} + \subgrad R_{k+1}}_*
  \\&\aleq{uses the triangle inequality}
  F(x_k)+ \tfrac{3}{2}L\eta^2
  +2\eta\norm{\nabla f(x_{k+1}) - m_{k+1}}_*
  - \eta\norm{\nabla f(x_{k+1}) + \subgrad R_{k+1}}_*,
\end{align*}
where \annotate.\qed

\subsubsection{Proof of \Cref{lem:momentum}}\label{proof:momentum}
We can express $m_{k+1} - \nabla f(x_{k})$ as follows using \cref{eq:momentum}:
\begin{align*}
  m_{k+1} - \nabla f(x_k)
  &=
  (1-\alpha)m_k + \alpha g(x_k;\xi_k) - \nabla f(x_k)
  \\&=
  (1-\alpha)(m_k - \nabla f(x_{k-1})) + \alpha(g(x_k;\xi_k) - \nabla f(x_k))
  \\&
  +(1-\alpha)(\nabla f(x_{k-1}) - \nabla f(x_k)).
\end{align*}
This implies the following for all $k \geq 0$:
\begin{align*}
  m_{k+1} - \nabla f(x_k)
  &=
  (1-\alpha)^{k+1}(m_0 - \nabla f(x_0))
  + \sum_{i=0}^{k-1}(1-\alpha)^{k-i}(\nabla f(x_{i}) - \nabla f(x_{i+1}))
  \\&
  +\sum_{i=0}^{k}\alpha(1-\alpha)^{k-i}(g(x_i,\xi_i) - \nabla f(x_i)).
\end{align*}
Using this, we can upper-bound $\E{\norm{m_{k+1} - \nabla f(x_k)}_*}$ for $k\geq 0$ as follows:
\begin{align*}
  \E{\norm{m_{k+1} - \nabla f(x_k)}_*}
  &\aleq{uses the triangle inequality}
  (1-\alpha)^{k+1}\E{\norm{m_0 - \nabla f(x_0)}_*}
  \\&
  + \sum_{i=0}^{k-1}(1-\alpha)^{k-i}\norm{\nabla f(x_{i}) - \nabla f(x_{i+1})}_*
  \\&
  +\E{\norm{\sum_{i=0}^{k}\alpha(1-\alpha)^{k-i}(g(x_i,\xi_i) - \nabla f(x_i))}_*}
  \\&\aleq{uses \Cref{eq:L} and \cref{eq:str}}
  (1-\alpha)^{k+1}\E{\norm{m_0 - \nabla f(x_0)}_*}
  + \sum_{i=0}^{k-1}(1-\alpha)^{k-i}L\eta
  \\&
  +\E{\norm{\sum_{i=0}^{k}\alpha(1-\alpha)^{k-i}(g(x_i,\xi_i) - \nabla f(x_i))}_*}
  \\&\aleq{uses \Cref{eq:norm}}
  (1-\alpha)^{k+1}\rho\E{\norme{m_0 - \nabla f(x_0)}}
  + \sum_{i=0}^{k-1}(1-\alpha)^{k-i}L\eta
  \\&
  +\rho\E{\norme{\sum_{i=0}^{k}\alpha(1-\alpha)^{k-i}(g(x_i,\xi_i) - \nabla f(x_i))}}
  \\&\aleq{uses Jensen's inequality}
  (1-\alpha)^{k+1}\rho\sqrt{\E{\sqne{m_0 - \nabla f(x_0)}}}
  + \sum_{i=0}^{k-1}(1-\alpha)^{k-i}L\eta
  \\&
  +\rho\sqrt{\E{\sqne{\sum_{i=0}^{k}\alpha(1-\alpha)^{k-i}(g(x_i,\xi_i) - \nabla f(x_i))}}}
  \\&\aleq{uses \Cref{eq:variance} and the fact that samples $\xi_i \sim \cD$ are i.i.d.}
  (1-\alpha)^{k+1}\rho\sigma
  + \sum_{i=0}^{k-1}(1-\alpha)^{k-i}L\eta
  +\alpha\rho\sigma\sqrt{\sum_{i=0}^{k}(1-\alpha)^{2(k-i)}}
  \\&\leq
  (1-\alpha)^{k+1}\rho\sigma + \frac{L\eta}{\alpha} + \sqrt{\alpha}\rho\sigma,
\end{align*}
where \annotate.\qed

\newpage

\subsection{Proof of \Cref{thm:tr_decay}}\label{proof:tr_decay}

We start with the following \Cref{lem:x}. The proof is available in \Cref{proof:x}.
\begin{lemma}\label{lem:x}
  Under the conditions of \Cref{thm:tr_decay}, let $x \in \sX$ be defined as follows:
  \begin{equation}\label{eq:x_def}
    x = \beta x^* + (1-\beta)x_k.
  \end{equation}
  Then, the following inequalities hold:
  \begin{equation}
    \norm{x-(1-\beta)x_k} \leq \eta,
    \quad
    \norm{x-x_k} \leq 2\eta,
    \quad
    \norm{x-x_{k+1}} \leq 2\eta,
    \quad
    \norm{x_{k+1}-x_k} \leq 2\eta.
  \end{equation}
\end{lemma}

Now, we are ready to prove \Cref{thm:tr_decay}. We can upper-bound $F(x_{k+1})$ as follows:
\begin{align*}
  F(x_{k+1})
  &\aeq{use the definition of function $F(x)$ in \cref{eq:main}}
  f(x_{k+1}) + R(x_{k+1})
  \\&\aleq{use \Cref{eq:L}}
  f(x_k) + \<\nabla f(x_k),x_{k+1} - x_k> + \tfrac{1}{2}L\sqn{x_{k+1} - x_k} + R(x_{k+1})
  \\&\aleq{use \Cref{lem:x}}
  f(x_k) + \<\nabla f(x_k),x_{k+1} - x_k> + R(x_{k+1}) + 2L\eta^2
  \\&\aleq{uses \cref{eq:tr_decay} and the inequality $\norm{x - (1-\beta)x_k}\leq \eta$, which is implied by \Cref{lem:x}}
  f(x_k) + \<\nabla f(x_k),x - x_k> + R(x) + 2L\eta^2
  \\&\aleq{use \Cref{eq:L}}
  f(x) + \tfrac{1}{2}L\sqn{x-x_k} + R(x) + 2L\eta^2
  \\&\aleq{use \Cref{lem:x}}
  f(x) + R(x) + 4L\eta^2
  \\&\aleq{uses \cref{eq:star_cvx} and the convexity if function $R(x)$}
  \beta (f(x^*) + R(x^*)) + (1-\beta) (f(x_k) + R(x_k)) + 4L\eta^2
  \\&\aeq{use the definition of function $F(x)$ in \cref{eq:main}}
  \beta F(x^*) + (1-\beta) F(x_k) + 4L\eta^2,
\end{align*}
where \annotate. After rearranging, we obtain
\begin{align*}
  F(x_{k+1}) - F(x^*) \leq (1-\beta)(F(x_k) - F(x^*)) + 4L\eta^2,
\end{align*}
which implies the following inequality:
\begin{align*}
  F(x_K) - F(x^*) \leq (1-\beta)^K(F(x_0) - F(x^*)) + \frac{4L\eta^2}{\beta}.
\end{align*}
\qed

\subsubsection{Proof of \Cref{lem:x}}\label{proof:x}
First, we can show that $\beta\norm{x_k} \leq \eta$ by induction. Indeed $\beta\norm{x_0} \leq \eta$ due to \cref{eq:eta_beta}, and for all $k \in \{0,1,\ldots,K-1\}$, can obtain the following:
\begin{align*}
  \beta\norm{x_{k+1}}
  \aleq{uses the triangle inequality}
  \beta\norm{x_{k+1} - (1-\beta)x_k} + (1-\beta)\beta\norm{x_k}
  \aleq{uses \cref{eq:tr_decay} and the induction hypothesis}
  \eta\beta + (1-\beta)\eta
  \leq \eta,
\end{align*}
where \annotate. Hence, we can prove the desired inequalities as follows:
\begin{align*}
  \norm{x - (1-\beta)x_k}
  &\aeq{use the definition of $x$ in \cref{eq:x_def}}
  \beta\norm{x^*}
  \aleq{uses \cref{eq:eta_beta}}
  \eta,
  \\
  \norm{x - x_k}
  &\aeq{use the definition of $x$ in \cref{eq:x_def}}
  \beta\norm{x^* - x_k}
  \aleq{use the triangle inequality}
  \beta\norm{x^*}+\beta\norm{x_k}
  \aleq{uses \cref{eq:eta_beta} and the previously obtained inequality $\beta\norm{x_k}\leq \eta$}
  2\eta,
  \\
  \norm{x - x_{k+1}}
  &\aleq{use the triangle inequality}
  \norm{x - (1-\beta)x_k} + \norm{x_{k+1} - (1-\beta)x_k}
  \aleq{uses \cref{eq:tr_decay} and the previously obtained inequality $\norm{x - (1-\beta)x_k}\leq \eta$}
  2\eta,
  \\
  \norm{x_{k+1} - x_k}
  &\aleq{use the triangle inequality}
  \norm{x_{k+1} - (1-\beta)x_k} + \beta\norm{x_k}
  \aleq{uses \cref{eq:tr_decay} and the previously obtained inequality $\beta\norm{x_k}\leq \eta$}
  2\eta,
\end{align*}
where \annotate.\qed

\newpage

\subsection{Proof of \Cref{thm:str_decay}}\label{proof:str_decay}

In this proof, we are going to use \Cref{lem:x}, which is valid not only for the iterations~\eqref{eq:tr_decay}, but also for \Cref{alg:decay}. We also obtain the following \Cref{lem:momentum_decay}. The proof is almost identical to the proof of \Cref{lem:momentum} in \Cref{proof:momentum}, with the only difference being that we use the inequality $\norm{x_{k+1} - x_k} \leq 2\eta$, which holds due to \Cref{lem:x}.
\begin{lemma}\label{lem:momentum_decay}
  Let \Cref{eq:L,eq:variance,eq:norm} hold, and let $x_0 \in \dom R$ and $m_0 = g(x_0,\xi_0)$. Then the iterations of \Cref{alg:decay} satisfy the following inequality for $k \geq 0$:
  \begin{equation}
    \E{\norm{m_{k+1} - \nabla f(x_{k})}_*} \leq  (1-\alpha)^{k+1}\rho\sigma + \sqrt{\alpha}\rho\sigma + \frac{2L\eta}{\alpha}.
  \end{equation}
\end{lemma}

Now, we are ready to prove \Cref{thm:str_decay}. We can upper-bound $F(x_{k+1})$ as follows:
\begin{align*}
  F(x_{k+1})
  &\aeq{use the definition of function $F(x)$ in \cref{eq:main}}
  f(x_{k+1}) + R(x_{k+1})
  \\&\aleq{use \Cref{eq:L}}
  f(x_k) + \<\nabla f(x_k),x_{k+1} - x_k> + \tfrac{1}{2}L\sqn{x_{k+1} - x_k} + R(x_{k+1})
  \\&\aleq{use \Cref{lem:x}}
  f(x_k) + \<\nabla f(x_k) - m_{k+1},x_{k+1} - x_k> + 2L\eta^2 + \<m_{k+1},x_{k+1} - x_k> +R(x_{k+1})
  \\&\aleq{uses \cref{eq:str_decay} and the inequality $\norm{x - (1-\beta)x_k}\leq \eta$, which is implied by \Cref{lem:x}}
  f(x_k) + \<\nabla f(x_k) - m_{k+1},x_{k+1} - x_k> + 2L\eta^2 + \<m_{k+1},x - x_k> +R(x)
  \\&=
  f(x_k) + \<\nabla f(x_k),x - x_k>  + 2L\eta^2 + R(x)
  +\<m_{k+1} - \nabla f(x_k),x - x_{k+1}>
  \\&\aleq{uses the definition of the dual norm}
  f(x_k) + \<\nabla f(x_k),x - x_k>  + 2L\eta^2 + R(x)
  +\norm{x - x_{k+1}}\norm{m_{k+1} - \nabla f(x_k)}_*
  \\&\aleq{use \Cref{lem:x}}
  f(x_k) + \<\nabla f(x_k),x - x_k>  + 2L\eta^2 + R(x)
  +2\eta\norm{m_{k+1} - \nabla f(x_k)}_*
  \\&\aleq{use \Cref{eq:L}}
  f(x) + \tfrac{1}{2}L\sqn{x-x_k}  + 2L\eta^2 + R(x)
  +2\eta\norm{m_{k+1} - \nabla f(x_k)}_*
  \\&\aleq{use \Cref{lem:x}}
  f(x) + 4L\eta^2 + R(x)
  +2\eta\norm{m_{k+1} - \nabla f(x_k)}_*
  \\&\aleq{uses \cref{eq:star_cvx} and the convexity if function $R(x)$}
  \beta (f(x^*) + R(x^*)) + (1-\beta) (f(x_k) + R(x_k)) + 4L\eta^2
  +2\eta\norm{m_{k+1} - \nabla f(x_k)}_*
  \\&\aeq{use the definition of function $F(x)$ in \cref{eq:main}}
  \beta F(x^*) + (1-\beta) F(x_k) + 4L\eta^2
  +2\eta\norm{m_{k+1} - \nabla f(x_k)}_*,
\end{align*}
where \annotate. After rearranging, taking the expectation, and using \Cref{lem:momentum_decay}, we get
\begin{align*}
  \E{F(x_{k+1}) - F(x^*)}
  &\leq
  (1-\beta) \E{F(x_k) - F(x^*)}
  +2\eta\rho\sigma(1-\alpha)^k + 2\eta\rho\sigma\sqrt{\alpha}
  \\&
  +4L\eta^2 + \frac{4L\eta^2}{\alpha},
\end{align*}
which implies the following inequality:
\begin{align*}
  \E{F(x_K) - F(x^*)} \leq (1-\beta)^K (F(x_0) - F(x^*)) + 2\eta\rho\sigma\left(\frac{1}{\alpha} + \frac{\sqrt{\alpha}}{\beta}\right) + \frac{4L\eta^2}{\beta}\left(1 + \frac{1}{\alpha}\right).
\end{align*}
\qed

\newpage

\subsection{Proof of \Cref{thm:hess_nonconvex}}\label{proof:hess_nonconvex}

In this proof, we are going to use \Cref{lem:descent}, whis is valid for the iterations of \Cref{alg:hess} as long as $\beta = 0$. We also obtain the following \Cref{lem:momentum_hess}, which describes the dynamics of the momentum $m_{k}$ under the second-order smoothness. Similarly to the proofs of \Cref{lem:momentum,eq:momentum_decay}, it can be seen as an adaptation of the proof by \citet{cutkosky2020momentum} to the non-Euclidean setting and is available in \Cref{proof:momentum_hess}.
\begin{lemma}\label{lem:momentum_hess}
  Let \Cref{eq:variance,eq:norm,eq:H} hold, and let $x_0 \in \dom R$ and $m_0 = g(x_0,\xi_0)$. Then the iterations of \Cref{alg:hess} satisfy the following inequality for $k \geq 0$:
  \begin{equation}
    \E{\norm{m_{k+1} - \nabla f(x_k)}_*} \leq (1-\alpha)^{k+1}\rho\sigma + \sqrt{\alpha}\rho\sigma
    +\frac{H\eta^2}{2\alpha^2}.
  \end{equation}
\end{lemma}
Now, we are ready to prove \Cref{thm:hess_nonconvex}. Using \Cref{lem:descent}, we obtain the following inequality:
\begin{align*}
  \min_{k=1,\ldots,K}\norm{\nabla f(x_k) + \subgrad R_k}_*
  &\leq
  \frac{F(x_0) - \inf_x F(x)}{\eta K} + \frac{3L\eta}{2} + \frac{2}{K}\sum_{k=1}^{K}\norm{\nabla f(x_k) - m_k}_*
  \\&\aleq{uses the triangle inequality and \Cref{eq:L}}
  \frac{F(x_0) - \inf_x F(x)}{\eta K} + \frac{7L\eta}{2} + \frac{2}{K}\sum_{k=0}^{K-1}\norm{\nabla f(x_k) - m_{k+1}}_*,
\end{align*}
where \annotate.
Using \Cref{lem:momentum_hess}, we obtain the following:
\begin{align*}
  \E{\min_{k=1,\ldots,K}\norm{\nabla f(x_k) + \subgrad R_k}_*}
  \leq
  \frac{F(x_0) - \inf_x F(x)}{\eta K}
  +\frac{7L\eta}{2}
  +\frac{H\eta^2}{\alpha^2}
  +\frac{2\rho\sigma}{\alpha K}
  +2\sqrt{\alpha}\rho\sigma.
\end{align*}
\qed

\subsubsection{Proof of \Cref{lem:momentum_hess}}\label{proof:momentum_hess}

We can express $m_{k+1} - \nabla f(x_{k})$ as follows using \cref{eq:momentum_hess}:
\begin{align*}
  m_{k+1} - \nabla f(x_k)
  &=
  (1-\alpha)m_k + \alpha g(\ox_k;\xi_k) - \nabla f(x_k)
  \\&=
  (1-\alpha)(m_k - \nabla f(x_{k-1})) + \alpha(g(\ox_k;\xi_k) - \nabla f(\ox_k))
  \\&
  +\alpha\nabla f(\ox^k)+(1-\alpha)\nabla f(x_{k-1}) - \nabla f(x_k)
\end{align*}
This implies the following for all $k \geq 0$:
\begin{align*}
  m_{k+1} - \nabla f(x_k)
  &=
  (1-\alpha)^{k+1}(m_0 - \nabla f(x_0))
  +\sum_{i=0}^{k}\alpha(1-\alpha)^{k-i}(g(\ox_i,\xi_i) - \nabla f(\ox_i))
  \\&
  +\sum_{i=0}^{k-1}(1-\alpha)^{k-i-1}(\alpha\nabla f(\ox_{i+1}) + (1-\alpha)\nabla f(x_{i}) - \nabla f(x_{i+1})).
\end{align*}
Using this, we can upper-bound $\E{\norm{m_{k+1} - \nabla f(x_k)}_*}$ for $k\geq 0$ as follows:
\begin{align*}
  \E{\norm{m_{k+1} - \nabla f(x_k)}_*}
  &\aleq{uses the triangle inequality}
  (1-\alpha)^{k+1}\E{\norm{m_0 - \nabla f(x_0)}_*}
  \\&
  + \sum_{i=0}^{k-1}(1-\alpha)^{k-i-1}\norm{\alpha\nabla f(\ox_{i+1}) + (1-\alpha)\nabla f(x_{i}) - \nabla f(x_{i+1})}_*
  \\&
  +\E{\norm{\sum_{i=0}^{k}\alpha(1-\alpha)^{k-i}(g(\ox_i,\xi_i) - \nabla f(\ox_i))}_*}
  \\&\aleq{uses \Cref{eq:norm}}
  (1-\alpha)^{k+1}\rho\E{\norme{m_0 - \nabla f(x_0)}}
  \\&
  + \sum_{i=0}^{k-1}(1-\alpha)^{k-i-1}\norm{\alpha\nabla f(\ox_{i+1}) + (1-\alpha)\nabla f(x_{i}) - \nabla f(x_{i+1})}_*
  \\&
  +\rho\E{\norme{\sum_{i=0}^{k}\alpha(1-\alpha)^{k-i}(g(\ox_i,\xi_i) - \nabla f(\ox_i))}}
  \\&\aleq{uses Jensen's inequality}
  (1-\alpha)^{k+1}\rho\sqrt{\E{\sqne{m_0 - \nabla f(x_0)}}}
  \\&
  + \sum_{i=0}^{k-1}(1-\alpha)^{k-i-1}\norm{\alpha\nabla f(\ox_{i+1}) + (1-\alpha)\nabla f(x_{i}) - \nabla f(x_{i+1})}_*
  \\&
  +\rho\sqrt{\E{\sqne{\sum_{i=0}^{k}\alpha(1-\alpha)^{k-i}(g(\ox_i,\xi_i) - \nabla f(\ox_i))}}}
  \\&\aleq{uses \Cref{eq:variance} and the fact that samples $\xi_i \sim \cD$ are i.i.d.}
  (1-\alpha)^{k+1}\rho\sigma + \alpha\rho\sigma\sqrt{\sum_{i=0}^{k}(1-\alpha)^{2(k-i)}}
  \\&
  + \sum_{i=0}^{k-1}(1-\alpha)^{k-i-1}\norm{\alpha\nabla f(\ox_{i+1}) + (1-\alpha)\nabla f(x_{i}) - \nabla f(x_{i+1})}_*
  \\&\aleq{uses the triangle inequality and \Cref{eq:H}}
  (1-\alpha)^{k+1}\rho\sigma + \sqrt{\alpha}\rho\sigma
  \\&
  + \sum_{i=0}^{k-1}(1-\alpha)^{k-i-1}
  \norm{\alpha\nabla^2 f(x_{i+1})(\alpha \ox_{i+1} +  (1-\alpha)x_{i} - x_{i+1})}_*
  \\&
  + \sum_{i=0}^{k-1}\frac{H(1-\alpha)^{k-i-1}}{2}\left(\alpha\sqn{\ox_{i+1} - x_{i+1}} + (1-\alpha)\sqn{x_i - x_{i+1}}\right)
  \\&\aeq{uses \cref{eq:ox_hess}}
  (1-\alpha)^{k+1}\rho\sigma + \sqrt{\alpha}\rho\sigma
  +\sum_{i=0}^{k-1}\frac{H(1-\alpha)^{k-i}}{2\alpha}\sqn{x_i - x_{i+1}}
  \\&\aleq{uses \cref{eq:str_hess}}
  (1-\alpha)^{k+1}\rho\sigma + \sqrt{\alpha}\rho\sigma
  +\sum_{i=0}^{k-1}\frac{H\eta^2(1-\alpha)^{k-i}}{2\alpha}
  \\&\leq
  (1-\alpha)^{k+1}\rho\sigma + \sqrt{\alpha}\rho\sigma
  +\frac{H\eta^2}{2\alpha^2},
\end{align*}
where \annotate.\qed

\newpage

\subsection{Proof of \Cref{thm:hess_convex}}\label{proof:hess_convex}

In this proof, we are going to use \Cref{lem:x}, which are valid for the iterations of \Cref{alg:hess}. We also obtain the following \Cref{lem:momentum_hess_decay}. The proof is almost identical to the proof of \Cref{lem:momentum_hess} in \Cref{proof:momentum_hess}, with the only difference being that we use the inequality $\norm{x_{k+1} - x_k} \leq 2\eta$, which holds due to \Cref{lem:x}.
\begin{lemma}\label{lem:momentum_hess_decay}
  Let \Cref{eq:variance,eq:norm,eq:H} hold, and let $x_0 \in \dom R$ and $m_0 = g(x_0,\xi_0)$. Then the iterations of \Cref{alg:hess} satisfy the following inequality for $k \geq 0$:
  \begin{equation}
    \E{\norm{m_{k+1} - \nabla f(x_k)}_*} \leq (1-\alpha)^{k+1}\rho\sigma + \sqrt{\alpha}\rho\sigma
    +\frac{2H\eta^2}{\alpha^2}.
  \end{equation}
\end{lemma}
Now, we are ready to prove \Cref{thm:hess_convex}. Similarly to the proof of \Cref{thm:str_decay} in \Cref{proof:str_decay}, we can upper-bound $F(x_{k+1}) - F(x^*)$ as follows:
\begin{align*}
  F(x_{k+1}) - F(x^*)
  \leq
  (1-\beta) (F(x_k) - F(x^*)) + 4L\eta^2
  +2\eta\norm{m_{k+1} - \nabla f(x_k)}_*,
\end{align*}
Using \Cref{lem:momentum_hess}, we obtain the following inequality:
\begin{align*}
  \E{F(x_{k+1}) - F(x^*)}
  &\leq
  (1-\beta) \E{F(x_k) - F(x^*)}
  +2\eta\rho\sigma(1-\alpha)^{k} + 2\eta\rho\sigma\sqrt{\alpha}
  \\&
  +4L\eta^2 +\frac{4H\eta^3}{\alpha^2},
\end{align*}
which implies the following inequality:
\begin{align*}
  \E{F(x_K) - F(x^*)}
  &\leq
  (1-\beta)^K (F(x_0) - F(x^*))
  +2\eta\rho\sigma\left(\frac{1}{\alpha} + \frac{\sqrt{\alpha}}{\beta}\right)
  +\frac{4L\eta^2}{\beta} +\frac{4H\eta^3}{\alpha^2\beta}.
\end{align*}
\qed

\subsection{Proof of \Cref{thm:tr_D}}\label{proof:tr_D}
We start with the following inequality obtained in the proof of \Cref{thm:tr_nonconvex} in \Cref{proof:tr_nonconvex}:
\begin{equation}\label{aux:6}
  F(x_{k+1}) \leq F(x_k) - \eta\norm{\nabla f(x_{k+1}) + \subgrad R_{k+1}}_* + \tfrac{3}{2}L\eta^2.
\end{equation}
Using this inequality, we can upper-bound $F(x_{k+1})$ as follows:
\begin{align*}
  F(x_{k+1})
  &\aeq{use the definition of $F(x)$ in \cref{eq:main}}
  f(x_{k+1}) + R(x_{k+1})
  \\&\aleq{uses the convexity of function $R(x)$ and \Cref{eq:star_cvx}}
  f(x^*) + R(x^*) + \<\nabla f(x_{k+1}) + \subgrad R_{k+1}, x_{k+1} - x^*>
  \\&\aeq{use the definition of $F(x)$ in \cref{eq:main}}
  F(x^*) + \<\nabla f(x_{k+1}) + \subgrad R_{k+1}, x_{k+1} - x^*>
  \\&\aleq{uses the definition of the dual norm}
  F(x^*) + \norm{x_{k+1} - x^*}\norm{\nabla f(x_{k+1}) + \subgrad R_{k+1}}_*
  \\&\aleq{uses \Cref{eq:D} and the fact that $x_{k+1},x^* \in \dom R$}
  F(x^*) + D\norm{\nabla f(x_{k+1}) + \subgrad R_{k+1}}_*
  \\&\aleq{uses \cref{aux:6}}
  F(x^*) + (D/\eta)(F(x_k) - F(x_{k+1})) + \tfrac{3}{2}LD\eta,
\end{align*}
where \annotate. After rearranging, we obtain
\begin{align*}
  F(x_{k+1}) - F(x^*) \leq \frac{D}{\eta+D}\left(F(x_k) - F(x^*) + \frac{3L\eta^2}{2}\right),
\end{align*}
which implies the following inequality:
\begin{align*}
  F(x_K) - F(x^*)
  \leq
  \left(\frac{D}{\eta+D}\right)^K (F(x_0) - F(x^*)) + \frac{3LD\eta}{2}.
\end{align*}
\qed

\subsection{Proof of \Cref{thm:str_D}}\label{proof:str_D}
Similarly to the proof of \Cref{thm:tr_D} in \Cref{proof:tr_D}, we can obtain the following inequality:
\begin{equation}
  F(x_{k+1}) \leq F(x^*) + D\norm{\nabla f(x_{k+1}) + \subgrad R_{k+1}}_*.
\end{equation}
Combining this inequality with \Cref{lem:descent} gives the following:
\begin{align*}
  F(x_{k+1}) - F(x^*)
  &\leq
  \frac{D}{\eta + D} \left(F(x_k) - F(x^*) + 2\eta\norm{\nabla f(x_{k+1}) - m_{k+1}}_*
  + \frac{3L\eta^2}{2}\right)
  \\&\aleq{uses the triangle inequality, \Cref{eq:L}, and \cref{eq:str}}
  \frac{D}{\eta + D} \left(F(x_k) - F(x^*) + 2\eta\norm{\nabla f(x_k) - m_{k+1}}_*
  + \frac{7L\eta^2}{2}\right),
\end{align*}
where \annotate.
After taking the expectation and using \Cref{lem:momentum}, we obtain the following:
\begin{align*}
  \E{F(x_{k+1}) - F(x^*)}
  &\leq
  \left(\frac{D}{\eta + D}\right)
  \E{F(x_k) - F(x^*)}
  + \frac{2L\eta^2}{\alpha} + \frac{3L\eta^2}{2}
  \\&
  +2(1-\alpha)^{k+1}\eta\rho\sigma + 2\sqrt{\alpha}\eta\rho\sigma,
\end{align*}
which implies the following inequality:
\begin{align*}
  \E{F(x_K) - F(x^*)}
  \leq
  \left(\frac{D}{\eta + D}\right)^K(F(x_0) - F(x^*))
  +2\sqrt{\alpha}D\rho\sigma
  +\frac{2\eta\rho\sigma}{\alpha}
  +\frac{3LD\eta}{2}
  +\frac{2LD\eta}{\alpha}.
\end{align*}
\qed

\end{document}